\PassOptionsToPackage{table,x11names,dvipsnames,xcdraw}{xcolor}
\documentclass[11pt]{article}

\usepackage[preprint]{acl}

\usepackage{times}
\usepackage{latexsym}

\usepackage[T1]{fontenc}

\usepackage[utf8]{inputenc}

\usepackage{microtype}

\usepackage{inconsolata}

\usepackage{graphicx}

\usepackage{iftex}
\ifPDFTeX\usepackage{CJKutf8}\else\usepackage{xeCJK}\setCJKsansfont{Noto Sans CJK KR}\newenvironment{CJK}[2]{}{}\fi
\usepackage{tipa}
\usepackage{newtxmath}
\usepackage{amsmath}
\usepackage{makecell}
\usepackage{array, multirow}
\usepackage{float}
\usepackage{xcolor}
\usepackage[normalem]{ulem}
\useunder{\uline}{\ul}{}
\usepackage{booktabs}
\usepackage{rotating}
\usepackage{multirow}
\usepackage{nicematrix}
\usepackage{color, colortbl}
\usepackage{hhline}
\usepackage{array}
\usepackage{pifont}
\usepackage{subfigure}     

\definecolor{Gray}{gray}{0.9}

\usepackage{enumitem}
\usepackage[super]{nth}
\usepackage{amsfonts}

\usepackage{pmboxdraw}

\usepackage{appendix}

\usepackage[most]{tcolorbox}
\definecolor{findbg}{HTML}{F1F1EF}
\definecolor{findrule}{HTML}{8A8984}

\usepackage{cleveref}
\usepackage{hyperref}

\title{\textsc{KoAGE}: Do Large Language Models Know, Learn, and Coin \\ Korean Neologisms?}

\author{
  \textbf{SungHo Kim}\textsuperscript{1},
  \textbf{Eojin Jeon}\textsuperscript{1},
  \textbf{SangKeun Lee}\textsuperscript{1,2}
\\
  \textsuperscript{1}Department of Artificial Intelligence, Korea University, Seoul, South Korea \\
  \textsuperscript{2}Department of Computer Science and Engineering, Korea University, Seoul, South Korea \\
\\
  \texttt{\{sungho3268, skdlcm456, yalphy\}@korea.ac.kr}
}

\newcommand{\ko}[1]{#1}
\newsavebox{\fitbox}
\newcommand{\fittable}[2]{\sbox{\fitbox}{#2}\ifdim\wd\fitbox>#1 \resizebox{#1}{!}{\usebox{\fitbox}}\else\usebox{\fitbox}\fi}   
\newcommand{\citereports}{\citeauthor{nam2021sineo2020}, \citeyear{nam2021sineo2020}--\citeyear{nam2025sineo2024}}   
\nocite{nam2022sineo2021,nam2023sineo2022,nam2024sineo2023}
\newcommand{\gl}[2]{\ko{#1}\,{\footnotesize\textcolor{gray!70!black}{\textit{#2}}}}   
\newtcolorbox{promptbox}[1]{enhanced, breakable, colback=gray!6, colframe=gray!40, boxrule=0.4pt, arc=1.5pt, left=5pt, right=5pt, top=3pt, bottom=3pt, fontupper=\small, title={#1}, coltitle=black, colbacktitle=gray!15, fonttitle=\small\bfseries, before skip=4pt, after skip=6pt}
\newcommand{\finding}[2]{\begin{tcolorbox}[enhanced, colback=findbg, colframe=findbg, borderline west={2pt}{0pt}{findrule}, boxrule=0pt, arc=1pt, left=6pt, right=6pt, top=3pt, bottom=3pt, boxsep=0pt, before skip=6pt, after skip=4pt]\textbf{Finding #1.}\ #2\end{tcolorbox}}

\begin{document}
\nocite{nam2022sineo2021,nam2023sineo2022,nam2024sineo2023}
\begin{CJK}{UTF8}{mj}
\maketitle

\begin{abstract}
Korean neologisms arise through diverse mechanisms, a demanding test of language models beyond lexical memorization. To enable systematic evaluation, we introduce \textsc{NeoTrace}, an automated, rerunnable framework that collects and dates neologisms across genres. Its 4,271 entries support \textsc{KoAGE}, a cutoff-aware benchmark with three acquisition and two generation tasks---the latter, to our knowledge, the first to compare model coinages with attested neologisms under their original materials and community context. Across 21 models, lexical recall diverges from contextual inference: in a five-model temporal analysis, models define only 13--30\% of cutoff-year words, yet can infer meaning from one usage. However, structural understanding remains limited: even correct definitions barely improve formation analysis of pre-cutoff neologisms. Given word-building materials, single models often follow human formation mechanisms, whereas multi-agent coinage favors abbreviation and depends on interaction framing. \textsc{KoAGE} thus separates what models know, learn from use, and coin.\footnote{Our code is available \href{https://github.com/SungHo3268/KoAGE}{here}.}
\end{abstract}

\section{Introduction}

\begin{figure}[t!]
    \centering
    \includegraphics[width=\columnwidth]{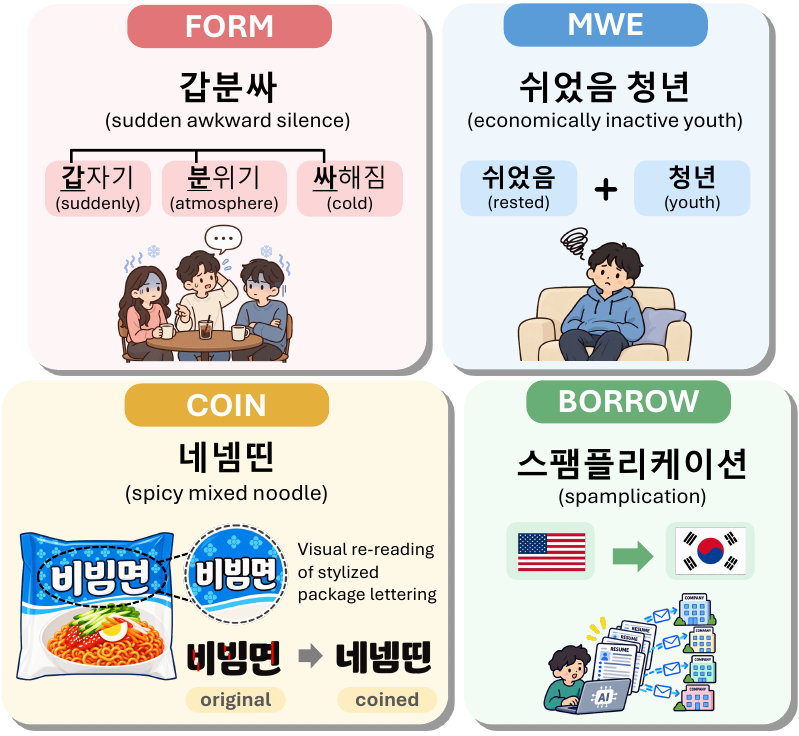}
    \caption{Overview of the taxonomy of the \textsc{KoAGE} lexicon: formation (FORM), multiword expression (MWE), borrowing (BORROW), and coinage (COIN).}
    \label{fig:neologism_overview}
\end{figure}

Language models are trained on a snapshot of language, while vocabulary continues to change. Korean offers a demanding test of adaptation because its neologisms exploit rich morphology and syllable-block writing~\citep{sohn1999korean}. Figure~\ref{fig:neologism_overview} illustrates this diversity: \gl{쉬었음 청년}{economically inactive youth} combines a nominalized verb form with a noun; \gl{갑분싸}{sudden awkward silence} retains one syllable from each of three constituent words~\citep{choi2022neologism}; and \gl{네넴띤}{\emph{bibimmyeon}} arises from the visual reinterpretation of stylized Hangul Jamo (subcharacter) lettering. Korean also draws on native, Sino-Korean, and Western lexical strata, whose elements combine productively in neologism formation~\citep{nam2022neologism}.

This diversity requires models to recover syllables deleted by abbreviation, combine morphemes across lexical strata, and recognize letter-level respellings.
Despite this heterogeneity in Korean word coinage, existing neologism benchmarks have largely focused on English~\citep{zheng2024neobench, deng2024newterm}. For Korean, KoNeoBench~\citep{lee2026koneobench} is the only dedicated benchmark, but it leaves critical problems open. Its items come from news articles alone, so the informal, community-driven coinages where word formation is most productive~\citep{bauer2001productivity,schmid2008newwords} are underrepresented; its results do not separate what a model memorized before its training cutoff from what it can infer after it; and it tests only the understanding of existing words, not whether a model can coin one. The last gap matters as AI agents engage in sustained interaction and human--agent collaboration~\citep{maharana2024evaluating,shao2026collaborativegym}, where using and coining new words becomes part of evolving linguistic practices.

Therefore, we develop an end-to-end framework, \textsc{NeoTrace}, for collecting and dating neologisms across genres. Deterministic extraction yields 3.9M candidate forms from a 12.8M-document corpus and search snippets, which are filtered using a temporal exclusion lexicon and evidence-grounded LLM verification. The resulting lexicon of 4,271 dated neologisms underlies \textsc{KoAGE}, a benchmark of three acquisition and two generation tasks comprising 5,930 test instances, with results reported separately on pre- and post-cutoff items for each model. Beyond understanding existing words, \textsc{KoAGE} evaluates coinage in two complementary settings: a single model constructs a new word from given morphological materials, while multiple agents interact in a reconstructed community context in which a real neologism emerged. We compare the resulting words with the attested human coinages to see where forms and formation strategies converge or diverge.

Across 21 models with cutoffs from 2021 to 2026, lexical recall and contextual inference diverge. In a five-model temporal analysis, models define only 13--30\% of words from their cutoff year, yet can infer meaning from one usage. This contextual success does not guarantee structural understanding: even correct definitions improve formation analysis of pre-cutoff words only marginally, while difficulty tracks formation type more closely than domain. Generation reveals another distinction. Given word-building materials, single models often follow human formation mechanisms, whereas multi-agent coinage favors abbreviation and depends on the interaction frame. Evaluating evolving vocabulary requires temporally aligning words and models, testing acquisition from use rather than recall alone, and comparing generated forms with attested human coinages.

\noindent Our contributions are as follows:
\begin{itemize}[leftmargin=1em, itemsep=0em]
    \item We introduce \textsc{NeoTrace}, an automatic and rerunnable framework for collecting and dating neologisms across genres under a unified Korean neologism taxonomy.
    \item We propose \textsc{KoAGE}, a benchmark to evaluate Korean neologism acquisition and generation capabilities, including both constrained coinage and socially situated multi-agent generation.
    \item We evaluate 21 proprietary and open-weight models on \textsc{KoAGE} and distill the results into six findings on how LLMs know, learn, and coin Korean neologisms.
\end{itemize}

\section{Related Work}

\paragraph{Neologism extraction.} Lexicographic monitoring systems find candidate new words by scanning dated text for forms absent from a reference lexicon, from the NeoCrawler for English \citep{kerremans2012neocrawler} and N\'eoveille for European languages \citep{cartier2017neoveille} to the annual Korean newspaper survey on which KoNeoBench builds (\citereports). \textsc{NeoTrace} reaches beyond the newspapers and single platforms of prior systems into messenger chat, video comments, and spoken dialogue, judges each candidate against the vocabulary known at its own date, and verifies with evidence automatically instead of by hand at a measured precision (Appendix~\ref{app:collection}).

\begin{table}[h]
\centering
\small
\setlength{\tabcolsep}{3.5pt}
\newcommand{\yes}{\textcolor{green!60!black}{\ding{51}}}\newcommand{\no}{\textcolor{red!80!black}{\ding{55}}}\newcommand{\half}{\textcolor{orange!80!black}{$\triangle$}}
\resizebox{\columnwidth}{!}{%
\begin{tabular}{l l r c c c}
    \toprule
    \textbf{Benchmark} & \textbf{Lang.} & \textbf{\#Neologisms} & \textbf{\#Genres} & \textbf{Cutoff split} & \textbf{Gen.} \\
    \midrule
    NEO-BENCH & EN & 2,505 & 3 & \half & \no \\
    SLANG & EN & 408 & 1 & \half & \no \\
    NewTerm & EN & 600 & 1 & \half & \no \\
    CNeo-Bench & ZH & 4,759 & 1 & \no & \no \\
    KoNeoBench & KO & 1,785 & 1 & \no & \no \\
    \rowcolor[HTML]{f0f1fa}
    \textsc{KoAGE} (ours) & KO & 4,271 & 6 & \yes & \yes \\
    \bottomrule
\end{tabular}}
\caption{Neologism benchmarks. \#Genres: counted among news, social media, messenger chat, video, spoken dialogue and lexicographic references; Cutoff split: results separated into pre- and post-cutoff items per model (\half\ = cutoffs shown or post-cutoff items only, no per-model split); Gen.: coinage evaluated.}
\label{tab:related}
\vspace{-8pt}
\end{table}

\paragraph{Neologism benchmarks.} Table~\ref{tab:related} compares the closest benchmarks. NEO-BENCH \citep{zheng2024neobench} tests perplexity, cloze, definition and translation on 2.5K English neologisms; SLANG \citep{mei2024slang} infers the meaning of post-cutoff Urban Dictionary entries; NewTerm \citep{deng2024newterm} refreshes 300 dictionary terms a year; CNeo-Bench \citep{zhao2026cneo} scales to 4.8K Chinese items with a validated LLM judge; KoNeoBench \citep{lee2026koneobench} covers 1,785 newspaper-sourced Korean items. WinoDict \citep{eisenschlos2023winodict} and CoLLEGe \citep{teehan2024college} probe acquisition with invented words, and naming games study conventions among LLM agents with artificial signals \citep{kouwenhoven2025shaping,ashery2025emergent}. Among these, only \textsc{KoAGE} compares model coinages with attested neologisms, both individually and within a simulated community, and \textsc{KoAGE} alone dates every item and splits results by each model's cutoff, keeping pre- and post-cutoff words of one lexicon side by side rather than limiting contamination by construction or refresh \citep{vu2024freshllms,wu2025antileak,white2025livebench,cheng2024dated}.

\begin{figure*}[t!]
    \centering
    \includegraphics[width=\textwidth]{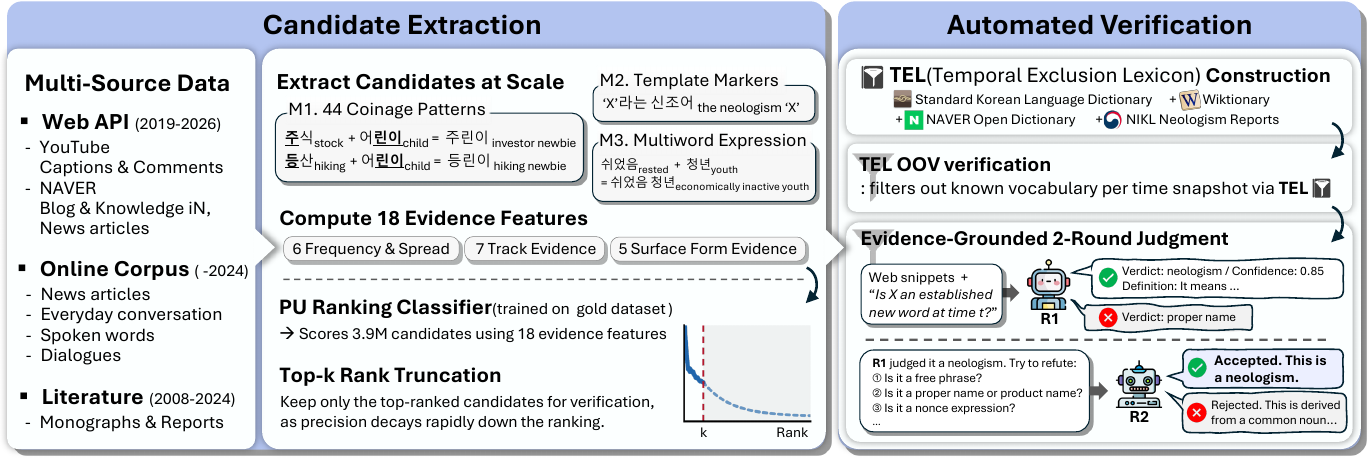}
    \caption{The \textsc{NeoTrace} framework: candidate extraction and automated verification over the three collection tracks. Validation is described in \S\ref{sec:validation}; details in Appendix~\ref{app:collection}.}
    \label{fig:main_methodology}
\end{figure*}

\section{\textsc{NeoTrace}}
\label{sec:neotrace}
We introduce \textsc{NeoTrace}, an end-to-end framework for collecting dated, verified neologisms at scale without expert curation. It comprises two stages (Figure~\ref{fig:main_methodology}): candidate extraction across genres (\S\ref{sec:extraction}) and verification against vocabulary attested at the time (\S\ref{sec:verification}). To assess the reliability of this approach, we compare its outputs with human judgments (\S\ref{sec:validation}). We focus on neologisms with novel forms, leaving semantic neologisms to lexical semantic change detection \citep{schlechtweg2020semeval,tahmasebi2021survey}, since a form's first attestation is observable whereas sense conventionalization is difficult to date reliably.

\subsection{Candidate Extraction}
\label{sec:extraction}
\paragraph{Multi-source data.}
Previous works draw mainly on newspapers or a single community platform (Table~\ref{tab:related}) and so under-represent informal coinages. We therefore collect through three tracks: an \emph{online corpus} of 12.8M documents (newspapers, video comments, daily conversations, community posts, and spoken dialogue; mostly 2017--2024); \emph{web API} retrieval of 207K NAVER\footnote{\href{https://www.naver.com}{NAVER} is the dominant Korean search platform.} search snippets from news, blogs, and Q\&A, plus captions and comments of 760 popular YouTube videos (2019--2026); and \emph{literature}, seven monographs \citep[\citereports;][]{choi2022neologism,nam2022neologism} and the annual neologism reports for 2008--2019 \citep{nikl2019collections}, used as a lexicographic reference.

\paragraph{Deterministic extraction.} We extract candidates from the corpus and the snippets using three types of surface evidence. First, we extract forms matching 44 productive coinage patterns derived from the lexicographic references, such as the splinter \ko{-린이} (from \gl{어린이}{child}). Second, we use nine metalinguistic markers that introduce unfamiliar words, such as \gl{`X'라는 신조어}{the neologism `X'}. Third, we target less overt coinages by extracting out-of-vocabulary forms after stripping particles and endings, as well as multiword expressions identified through part-of-speech patterns and association statistics, such as \gl{쉬었음}{rested}+\gl{청년}{youth}. Together, these extractors yield 3.9M candidates.

\begin{figure}[h]
    \centering
    \includegraphics[width=\columnwidth]{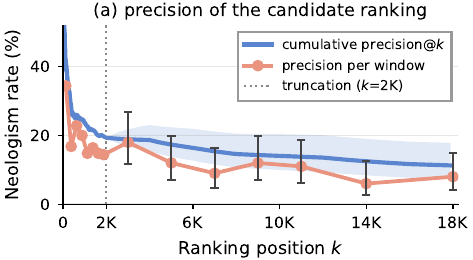}\\[6pt]
    \includegraphics[width=\columnwidth]{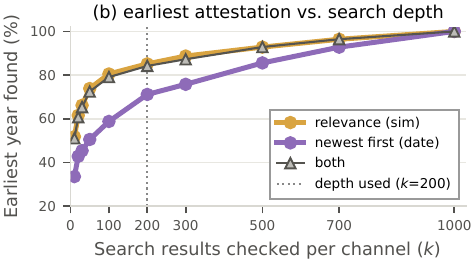}
    \caption{(a) Neologism rate by candidate rank: every candidate up to 2K verified, 100 sampled per window beyond. (b) Share of 200 sampled entries whose earliest sense-matched attestation lies within the top $k$ results per channel.}
    \label{fig:rankprec}
\end{figure}

\paragraph{Ranking and truncation.} Because manually verifying all 3.9M candidates is infeasible, we rank them by their similarity to catalogued neologisms in the lexicographic references (Appendix~\ref{app:collection}). Each candidate is represented by 18 features of distribution, surface form, and extraction evidence (Table~\ref{tab:features}). A classifier trained on catalogued neologisms as positive examples produces the ranking. We retain only the top-$k$ candidates for final verification, choosing $k=2{,}000$ at the knee of Figure~\ref{fig:rankprec}, where cumulative precision is 19\%.

\subsection{Automated Verification}
\label{sec:verification}
\paragraph{Judging relative to time.} A candidate that survives ranking may still be an old word, a name, or a one-off. To remove known vocabulary, we build a temporal exclusion lexicon (TEL) of 1.7M forms from three sources of established vocabulary: dictionary entries (the \emph{Standard Korean Language Dictionary} and Wiktionary), previously catalogued neologisms, and high-frequency forms of earlier corpora. The dictionary layers are static, while the corpus and prior-neologism layers use the candidate's year; candidates matching an applicable exclusion layer are dropped (Appendix~\ref{app:collection}).


\paragraph{LLM judgment.} Candidates that pass the TEL receive LLM evaluation. In the first pass, the judge receives an evidence packet rather than a bare word: dated usages, TEL status of the word and its parts, and yearly frequency. It answers the era-relative question, ``was this expression new in Korean around its first year?'', returning a neologism label, confidence score, definition, and cited evidence. A judge from another vendor rechecks low-confidence verdicts and a random sample of confident ones; disagreements are adjudicated. Rejections are kept as hard negatives for the contextual-choice task (\S\ref{sec:koage}).

\subsection{Validation}
\label{sec:validation}
\begin{figure}[h]
    \centering
    \includegraphics[width=\columnwidth]{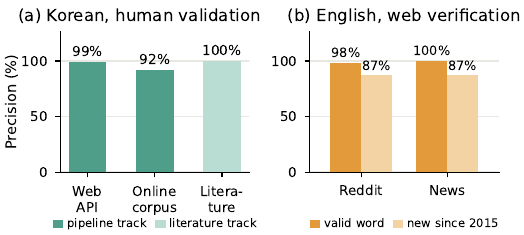}
    \caption{(a) Blind human validation of accepted Korean items. (b) Criterion-wise web verification of accepted items from Reddit and news.}
    \label{fig:validation}
    \vspace{-6pt}
\end{figure}

\paragraph{Against human judgment.}
\textsc{NeoTrace} admits or rejects candidates automatically; human judgment enters only to verify that output. To validate collection precision, we sampled 200 accepted items, 100 per automatic track, stratified by period and mixed with control negatives, and had an annotator blindly judge each as a neologism or not with its definition and usages in view (Figure~\ref{fig:validation}(a)). Overall precision reaches 95.5\%, against 19\% for the raw top-2k ranking.

\paragraph{Language-agnostic.} \textsc{NeoTrace} is language-agnostic except for three components: the normalizer, marker patterns, and morphological analyzer. To test transferability, we replaced these with English equivalents and re-ran \textsc{NeoTrace} on Reddit comments and online news. Of the top 2,000 candidates from each source, it accepted 210 and 160 neologisms, respectively; 98--100\% were valid words, and 87\% had emerged after 2015 according to web re-verification (Figure~\ref{fig:validation}(b); Appendix~\ref{app:english}).

\subsection{Neologism Lexicon}
\label{sec:taxonomy}
\textsc{NeoTrace} yields 4,271 verified neologisms, classified under a two-level taxonomy that follows Korean word-formation research \citep{choi2022neologism,nam2022neologism} and the categories of NEO-BENCH \citep{zheng2024neobench}, with four main categories (details in Appendix~\ref{app:taxonomy} and Appendix~\ref{app:composition}):
\begin{itemize}[nosep,leftmargin=1em]
    \item \textbf{FORM}: a word formed within Korean from existing material, native or foreign
    \item \textbf{MWE}: a new fixed multiword expression
    \item \textbf{BORROW}: a word imported whole from another language
    \item \textbf{COIN}: a word with no pre-existing material, made by sound or script play
\end{itemize}

\paragraph{Metadata.} Each entry includes a definition, usage examples, and a first-attestation year. Report definitions come from lexicographers, while pipeline definitions are generated by the judge from the evidence packet. Usage examples are drawn from the monographs, our dated corpus documents, or news retrieved through search; each is LLM-checked for intended-sense usage rather than definition. We take the first-attestation year from a dated source when available; otherwise, we use the earliest sense-matched attestation in our corpora or the top-200 relevance-ranked results, a depth chosen at the knee of Figure~\ref{fig:rankprec}(b). This date agrees with lexicographers' dates within one year for 90\% of checkable items.

\begin{figure*}[t!]
    \centering
    \includegraphics[width=\textwidth]{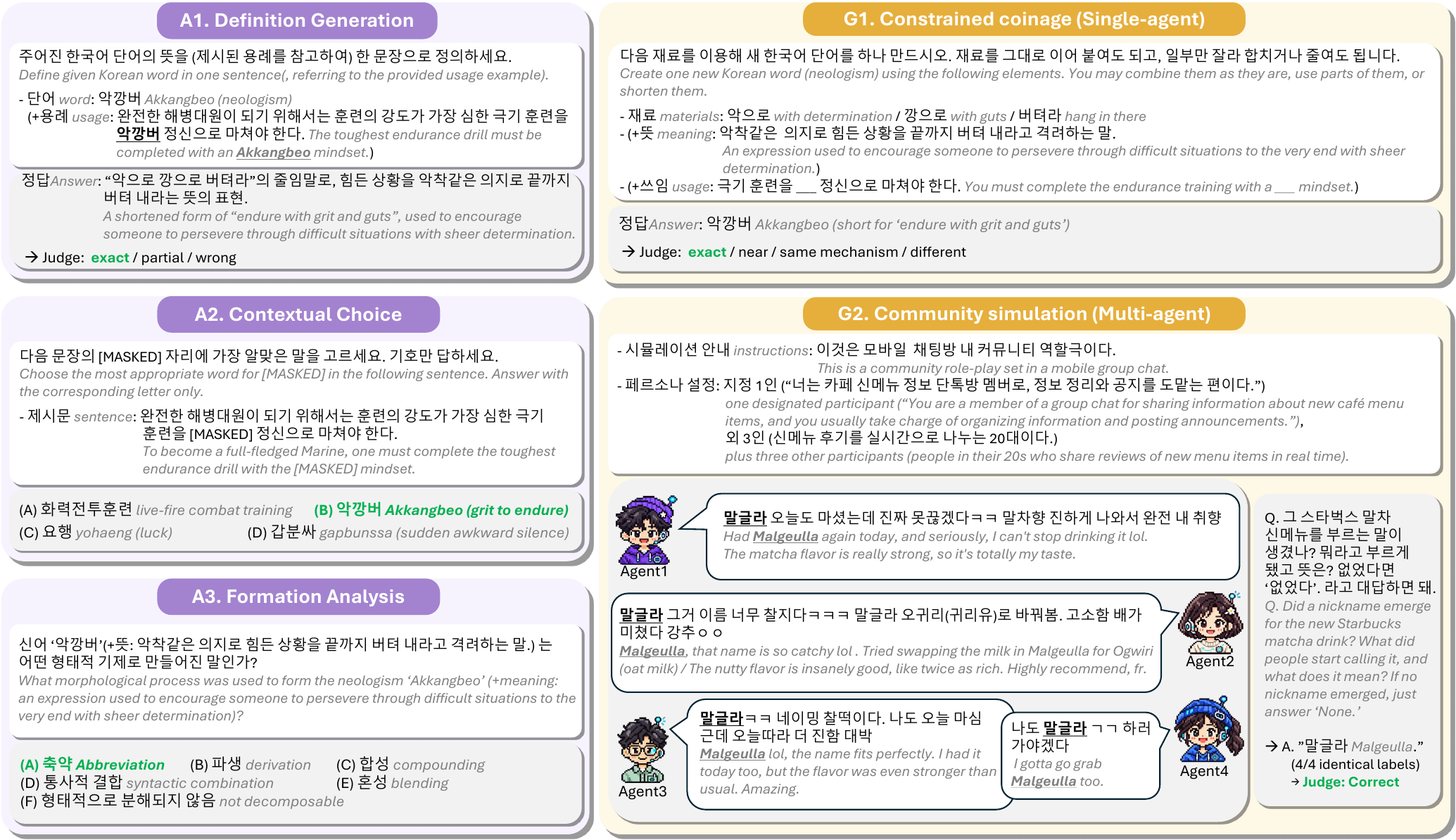}
    \caption{The \textsc{KoAGE} tasks with one example each: input and expected output for the three acquisition tasks (A1--A3) and the two generation tasks (G1--G2).}
    \label{fig:examples}
\end{figure*}

\section{\textsc{KoAGE}}
\label{sec:koage}

Using the neologisms collected above, we introduce \textsc{KoAGE}, a benchmark designed to evaluate models' ability to understand and generate Korean neologisms. \textsc{KoAGE} comprises 5,930 instances organized along two axes (Figure~\ref{fig:examples}): three \emph{acquisition} tasks (A1--A3) that assess neologism understanding and two \emph{generation} tasks (G1--G2) that assess neologism coinage. Each task draws on a relevant subset of the lexicon (Appendix~\ref{app:settings}).

\begin{table*}[t]
\centering\small
\setlength{\tabcolsep}{4pt}
\renewcommand{\arraystretch}{0.95}
\resizebox{\textwidth}{!}{%
\begin{tabular}{l c ccccc ccc ccc}
\toprule
& & \multicolumn{5}{c}{\textbf{Acquisition} (accuracy, \%; pre- / post-cutoff)} & \multicolumn{3}{c}{\textbf{G1 coinage} (exact, \%)} & \multicolumn{3}{c}{\textbf{G2 community} (\%)} \\
\cmidrule(lr){3-7} \cmidrule(lr){8-10} \cmidrule(lr){11-13}
\textbf{Model} & \textbf{Cutoff} & A1-W & A1-WU & A2 & A3-W & A3-WD & M & MD & MDC & Formed & Reprod. & Mech. \\
\midrule
\rowcolor{gray!15}\multicolumn{13}{l}{\textit{Proprietary (API)}} \\
GPT-3.5 Turbo & 2021-09 & 0.9 / 0.1 & 6.7 / 5.3 & 71.1 / 63.1 & 23.6 / 21.2 & 20.5 / 19.0 & 15.5 & 16.7 & 17.4 & 0.0 & 0.0 & 0.0 \\
GPT-4o & 2023-10 & 22.1 / 7.1 & 45.2 / 31.6 & 90.5 / 84.4 & 51.6 / 48.9 & 55.6 / 56.4 & 29.0 & 32.9 & 36.7 & 20.0 & 0.0 & 12.0 \\
GPT-5 & 2024-09 & 51.2 / 21.2 & 70.9 / 49.4 & 97.1 / 92.2 & 67.9 / 59.0 & 74.5 / 77.6 & 44.7 & 47.1 & 45.8 & 86.4 & 22.7 & 22.7 \\
GPT-5.4 mini & 2025-08 & 32.1 / 15.4 & 57.3 / 32.0 & 89.7 / 78.0 & 61.0 / 56.6 & 66.0 / 70.7 & 39.2 & 36.7 & 42.0 & 65.2 & 13.0 & 26.1 \\
GPT-5.6 sol & 2026-02 & 57.5 / 19.5 & 70.0 / 41.5 & 97.8 / 100.0 & 79.5 / 67.3 & 82.3 / 82.7 & 63.1 & 60.6 & 64.5 & 100.0 & 47.4 & 21.1 \\
Claude Sonnet 4.5 & 2025-07 & 30.0 / 10.8 & 52.7 / 28.6 & 90.9 / 86.0 & 61.4 / 55.8 & 65.6 / 71.2 & 41.4 & 40.7 & 49.4 & 95.2 & 42.9 & 9.5 \\
Claude Sonnet 5 & 2026-01 & 40.8 / 11.2 & 62.2 / 41.8 & 92.8 / 86.4 & 68.1 / 58.2 & 77.8 / 78.2 & 44.3 & 44.6 & 53.0 & 95.8 & 29.2 & 33.3 \\
Gemini 3 Flash & 2025-01 & 64.9 / 25.8 & 71.9 / 51.8 & 98.3 / 94.6 & 76.0 / 67.6 & 77.6 / 72.7 & 57.0 & 58.9 & 56.9 & 100.0 & 44.4 & 16.7 \\
Gemini 3.5 Flash-Lite & 2026-03 & 32.9 / 12.5 & 45.3 / 17.4 & 95.0 / 90.5 & 61.8 / 62.2 & 66.1 / 66.7 & 45.9 & 44.6 & 49.2 & 91.7 & 29.2 & 12.5 \\
Solar Pro 2 & 2025-03 & 20.4 / 2.1 & 42.9 / 26.9 & 89.3 / 78.8 & 44.3 / 42.5 & 48.7 / 50.4 & 26.0 & 24.8 & 28.2 & 66.7 & 4.2 & 25.0 \\
Solar Pro 4 & 2026-02 & 30.0 / 1.2 & 46.7 / 26.4 & 92.5 / 86.4 & 53.1 / 38.5 & 56.9 / 73.1 & 33.5 & 33.8 & 28.8 & 79.2 & 12.5 & 29.2 \\
HCX-003 & 2024-01$^{r}$ & 35.1 / 7.4 & 51.8 / 31.5 & 89.5 / 77.4 & 22.9 / 26.2 & 22.4 / 22.6 & 24.1 & 24.7 & 22.9 & 29.2 & 8.3 & 0.0 \\
HCX-007 & 2025-07$^{r}$ & 33.8 / 9.6 & 47.9 / 21.0 & 90.2 / 65.1 & 29.0 / 26.9 & 29.5 / 29.8 & 22.4 & 26.5 & 26.6 & 66.7 & 12.5 & 29.2 \\
\cmidrule(lr){1-13}
Mean (proprietary) & & 34.7 / 11.1 & 51.7 / 31.2 & 91.1 / 83.3 & 53.9 / 48.5 & 57.2 / 59.3 & 37.4 & 37.9 & 40.1 & 68.9 & 20.5 & 18.3 \\
\midrule
\rowcolor{gray!15}\multicolumn{13}{l}{\textit{Open-weight}} \\
Qwen2 7B & 2024-06$^{r}$ & 6.5 / 0.8 & 19.6 / 9.0 & 71.5 / 58.5 & 19.0 / 24.8 & 23.5 / 28.0 & 5.4 & 4.6 & 5.0 & 12.0 & 0.0 & 4.0 \\
Qwen2.5 7B & 2024-09$^{r}$ & 12.0 / 0.8 & 26.4 / 17.1 & 73.3 / 56.2 & 24.8 / 26.9 & 22.2 / 24.4 & 12.2 & 12.8 & 15.0 & 0.0 & 0.0 & 0.0 \\
Qwen3 8B & 2025-04$^{r}$ & 4.0 / 0.0 & 29.7 / 15.7 & 75.9 / 62.7 & 29.9 / 32.8 & 27.0 / 36.8 & 12.1 & 11.5 & 12.3 & 32.0 & 4.0 & 16.0 \\
Qwen3.5 9B & 2026-02$^{r}$ & 6.7 / 0.0 & 31.7 / 9.4 & 81.5 / 81.8 & 28.6 / 26.9 & 31.1 / 42.3 & 13.3 & 13.4 & 13.6 & 44.0 & 8.0 & 12.0 \\
EXAONE 3.0 7.8B & 2024-08$^{r}$ & 22.5 / 3.3 & 35.5 / 19.5 & 71.5 / 45.3 & 23.5 / 24.2 & 21.0 / 27.4 & 8.4 & 8.6 & 6.8 & 16.0 & 0.0 & 8.0 \\
EXAONE 3.5 7.8B & 2024-12$^{r}$ & 18.4 / 2.6 & 37.3 / 20.1 & 76.5 / 60.3 & 20.8 / 22.2 & 16.5 / 24.8 & 7.3 & 6.9 & 7.2 & 60.0 & 8.0 & 16.0 \\
EXAONE 4.0 32B & 2025-07$^{r}$ & 1.2 / 0.0 & 29.2 / 9.5 & 81.2 / 58.1 & 26.9 / 26.9 & 26.7 / 29.8 & 13.5 & 13.7 & 12.7 & 4.0 & 0.0 & 0.0 \\
EXAONE 4.5 33B & 2024-12 & 17.8 / 1.7 & 42.2 / 22.0 & 80.9 / 55.6 & 38.5 / 32.0 & 32.5 / 35.3 & 16.7 & 17.0 & 15.5 & 24.0 & 4.0 & 4.0 \\
\cmidrule(lr){1-13}
Mean (open-weight) & & 11.1 / 1.1 & 31.4 / 15.3 & 76.5 / 59.8 & 26.5 / 27.1 & 25.1 / 31.1 & 11.1 & 11.1 & 11.0 & 24.0 & 3.0 & 7.5 \\
\bottomrule
\end{tabular}}
\caption{\textsc{KoAGE} results for 21 models across acquisition and generation tasks. Acquisition cells report accuracy on items dated before (pre-cutoff) and after (post-cutoff) each model's training cutoff; item counts are provided in Table~\ref{tab:seen}. Cutoff denotes the documented training cutoff; $^{r}$ indicates the release month when no cutoff is reported.}
\label{tab:main}
\vspace{-5pt}
\end{table*}

\paragraph{A1 Definition generation.} Given the word alone (W) or with one attested usage sentence (WU), produce a one-sentence definition; a judge scores exact/partial/wrong against the reference. W measures recall; WU measures acquisition; their difference on post-cutoff items is the \emph{acquisition gain}.
\paragraph{A2 Contextual choice.} Fill a masked sentence from four choices: the answer, a neologism of the same formation type, a hard negative from the collection funnel, and a near-synonym that clashes with the context.
\paragraph{A3 Formation analysis.} Choose the formation mechanism of a FORM entry among the five mechanisms and `not decomposable', from the word alone (W) or with its definition (WD), which separates analysis from word knowledge.
\paragraph{G1 Constrained coinage.} Given the morphological materials (M), plus the concept rewritten without any hint of how the word is built (MD), plus a masked attested context (MDC), produce the word. We score exact and near form match and mechanism match against the gold word (Appendix~\ref{app:judge}); G1 thus measures constrained reconstruction, not free invention.
\paragraph{G2 Community simulation.} For neologisms with verified origins we reconstruct the emergence context: a community, a concept description without the target word, personas, and a scenario requiring a post conveying the situation to outsiders. Four agents of one backbone interact over twelve episodes, with no instruction to coin a word. Each agent is then asked independently whether a name emerged; a judge from another vendor finds the name shared by at least two agents and relates it to the attested coinage as exact, near, same mechanism, or different.

To our knowledge, G1 and G2 are the first tasks to compare model coinages with attested neologisms under their original materials and setting: does the model make the same formation choices as the speakers who coined it, alone (G1) and in a simulated community (G2)? Both use only words the model could not define in isolation (unknown in A1-W), limiting explicit lexical recall (Appendix~\ref{app:knownword}).

\section{Experimental Setup}
\label{sec:setup}
\paragraph{Models.} We evaluate 21 models spanning training cutoffs from 2021 to 2026: 13 proprietary models served through APIs (GPT-3.5 Turbo, GPT-4o, GPT-5, GPT-5.4 mini, GPT-5.6 sol; Claude Sonnet 4.5 and 5; Gemini 3 Flash and 3.5 Flash-Lite; Solar Pro 2 and Pro 4; HyperCLOVA X 003 and 007) and eight open-weight models run locally (EXAONE 3.0--4.5, Qwen 2--3.5). Details and references are in Table~\ref{tab:models}. Where a vendor publishes no cutoff, the release month is used, which makes the post-cutoff sets of those models conservative.

\paragraph{Judges and metrics.} Prompts are identical across models (Appendix~\ref{app:prompts}), options are shuffled with a fixed seed, and masks are fixed-length so word length is not leaked. A1 is scored by an LLM judge from a vendor other than the evaluated model as exact, partial or wrong against the reference definition; we report the exact rate, with abstentions counted as wrong, and the judge is validated against answers, a second judge and a human rater (81\% agreement, $\kappa$ 0.72; Appendix~\ref{app:judge}). A2 and A3 are scored by accuracy, G1 by exact form match with the attested word, and G2 by a judge of the exit interviews validated against a human rater in the appendix. Each result is a single run with item-level 95\% confidence intervals in the appendix; G2 uses card-level Wilson score intervals. Decoding settings are in Appendix~\ref{app:settings}.

\section{Results and Key Findings}
\label{sec:results}
We utilize \textsc{KoAGE} tasks to evaluate the ability of various LLMs to adapt to neologisms. Table~\ref{tab:main} gives the main results on \textsc{KoAGE} and the following are our key findings:

\finding{1}{Models rarely define a post-cutoff neologism, yet from one usage they can usually place it in context and often define it.}

From the word alone, proprietary models define 34.7\% of pre-cutoff and 11.1\% of post-cutoff items (A1-W). A single usage sentence adds about 20 points, but still leaves two-thirds of post-cutoff words undefined. By contrast, masked-usage choice (A2) is far easier for every model (91.1 / 83.3\% for proprietary models), both pre- and post-cutoff. Selection is weaker evidence than definition: A2 stays at 85.9\% on words a model could not define, while one usage lifts their exact definitions to 39.8\% pre- and 25.8\% post-cutoff (Appendix~\ref{app:robust}). This contextual inference underlies adaptation to language change: humans likewise acquire much vocabulary incidentally from context rather than definitions \citep{nagy1985context}, and \citet{eisenschlos2023winodict} probe the same capacity in LLMs.

\finding{2}{Adaptation to Korean neologisms is still weak and bounded by pretraining: every model knows little of the words from its own cutoff year.}

\begin{figure}[h]
    \centering
    \includegraphics[width=.97\columnwidth]{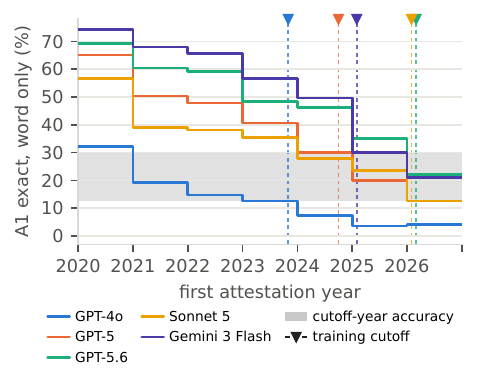}
    \caption{Word-only definition accuracy by year of first attestation. The shaded band spans the accuracy of the five models on words from their own cutoff year.}
    \label{fig:timeline}
\end{figure}

Figure~\ref{fig:timeline} shows that every model declines on newer words and that later cutoffs lift the whole curve. The striking regularity is the band: each model in the figure defines only 13--30\% of the words from its own cutoff year, whatever its overall ability. What a model has not been trained on stays largely unknown to it, however strong its general Korean. High scores on Korean benchmarks thus certify the conventions learned so far; keeping up with the language requires continual learning of the words that appear after the cutoff. Web search narrows this gap without closing it (Appendix~\ref{app:acq_anal}).

\finding{3}{A model's prior analysis of a familiar word resists even the correct definition.}
A3 asks the model to identify the formation mechanism of a FORM neologism, given either the word alone (W) or the word with its definition (WD). For pre-cutoff items, providing the definition improves accuracy by only 3.3 percentage points, compared with 10.8 points for post-cutoff items. With the definition, post-cutoff accuracy even surpasses pre-cutoff accuracy. This reversal (57.2\% vs.\ 59.3\%) suggests that a model's prior analysis of a familiar word is not overridden by the correct definition, or even by its constituents (Appendix~\ref{app:robust}).

\finding{4}{Domain differences follow formation types that make up each domain, and dissolve once the task adds a usage or a definition.}

\begin{figure}[h]
    \centering
    \includegraphics[width=\columnwidth]{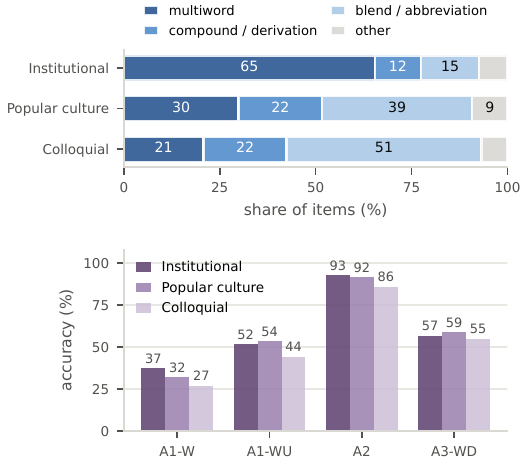}
    \caption{Acquisition performance by domain. Top: distribution of A1 items in each domain by formation type. Bottom: accuracy across the four acquisition regimes for 13 proprietary models.}
    \label{fig:domain}
    \vspace{-9pt}
\end{figure}

Figure~\ref{fig:domain} splits the lexicon into three domains (Appendix~\ref{app:domain}). They differ mainly in form: two-thirds of institutional words are multiword noun sequences whose parts supply the meaning, half of colloquial words are blends and abbreviations that delete it. Where the word form is all the model has (A1-W), accuracy follows this composition, falling from 37\% to 32\% to 27\% in order, and the masked-usage choice (A2) keeps the same order (93 to 86\%). Where the task supplies more information, a usage (A1-WU) or the definition (A3-WD), the order breaks, because context pays most where form says least: a usage adds 13 points to institutional words but 20--21 to the other two domains. Colloquial words, half of them blends and abbreviations, are hardest throughout, consistent with how poorly models read meaning off a form that has deleted its parts.

\finding{5}{With materials, a single model coins like humans; without, it diverges like a community, which alone drifts toward abbreviation.}

\begin{figure}[t!]
    \centering
    \includegraphics[width=\columnwidth]{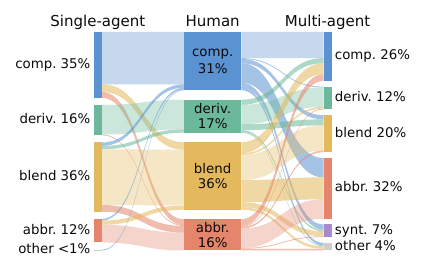}
    \caption{Formation mechanism of the same word by the single agent given its materials (G1-M), by the humans, and by the multi-agent community (G2). Pooled over 225 formed sessions of 21 backbones (Appendix~\ref{app:g2abl}).}
    \label{fig:sankey}
    \vspace{-9pt}
\end{figure}

Unlike earlier neologism benchmarks, \textsc{KoAGE} also evaluates coinage. In G1, the exact form match is low (Table~\ref{tab:main}), so a single model rarely produces the human word; Figure~\ref{fig:sankey} shows why this understates the resemblance: the single agent uses the human mechanism in 82\% of cases (Appendix~\ref{app:robust}). It combines the materials as people do and differs in length, order, position, and the language of the parts (native, Sino-Korean, English, or mixed). In G2, we simulate the community in which a word emerged and ask whether a neologism forms and how it compares with the human one. Here the shared word follows the human mechanism in only 48\% of cases: where the single agent and the human favor compounds and blends, communities shift to abbreviation (32\% of names, against 16\% of the human coinages and 12\% of the single-agent ones). This may reflect the pressure to shorten what is said often, which language users apply when they communicate \citep{piantadosi2011word,kanwal2017zipf}, and the method sentence of the protocol also plays a part (Finding~6). Because the simulation needs words whose origin is traceable, it covers 25 cards, and its findings call for deeper analysis.

\finding{6}{A concrete social context and a neutral reminder that words can be formed elicit coinage effectively, but they also shape the words that emerge.}

\begin{figure}[h]
    \centering
    \includegraphics[width=.97\columnwidth]{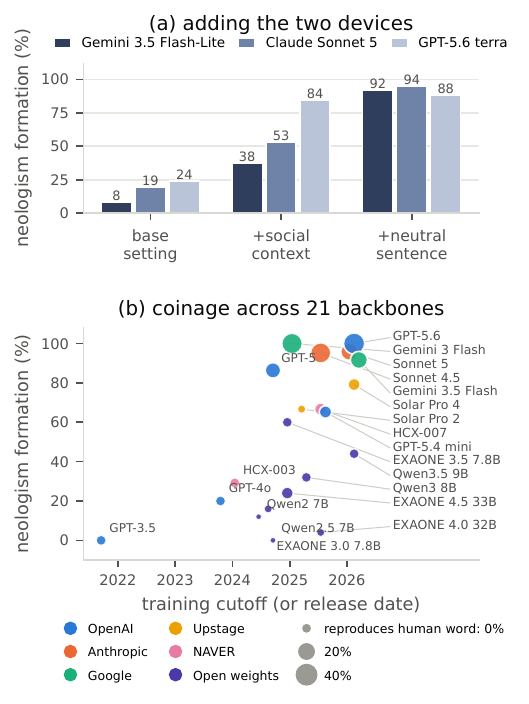}
    \caption{(a) Adding the two protocol devices cumulatively on three backbones. (b) Formation under the full protocol vs.\ cutoff (bubble size: share of sessions reproducing the human word).}
    \label{fig:g2}
    \vspace{-3pt}
\end{figure}

Left to converse, communities rarely coin (Figure~\ref{fig:g2}a: 8--24\% of sessions). We therefore use two elicitation devices: a realistic social context (personas and verbal culture) and a neutral sentence listing Korean word-formation methods without favoring any or mentioning the target. The social context raises formation to 38--84\%, and the neutral sentence to 88--94\%. The devices are not free of side effects: the method sentence steers the form of the words, with abbreviations at 44\% under a shortening-only wording and 32\% under the neutral one, against 18\% of the human coinages in the same sessions.
Across backbones (Figure~\ref{fig:g2}b), formation under the full protocol tracks the generation, from 0\% (GPT-3.5 Turbo) through 20--29\% (GPT-4o, HyperCLOVA X 003) to 65--100\% for the 2025--2026 proprietary models, while open-weight models spread from 0 to 60\% with no trend that two model sizes can establish. Community coinage is a capability of the recent generation, but what is measured depends on the frame that elicits it.
Tests with mixed-vendor communities appear in Appendix~\ref{app:mixed}.

\section{Conclusion}
\label{sec:discussion}
We introduced \textsc{NeoTrace}, a rerunnable framework for collecting and dating Korean neologisms, and \textsc{KoAGE}, a benchmark spanning three acquisition and two generation tasks. Across 21 models, we find that adapting to new vocabulary is not a single ability. Lexical recall declines near the training frontier, while models can infer the meanings of new words from context. Yet contextual inference does not ensure structural understanding: formation analyses of pre-cutoff words remain unreliable even with correct definitions, and acquisition difficulty tracks formation type more closely than domain. Human-like formation choices in constrained coinage depend on the materials supplied; in simulated communities abbreviation becomes more common and outcomes depend on interaction framing. These findings motivate evaluations that align words with model cutoffs, distinguish inference from recall, and compare generated words with attested human coinages while accounting for the conditions that elicit them. We release the framework, dated lexicon, tasks, and simulation harness to support such evaluation as linguistic conventions evolve.

\section*{Limitations}
Although \textsc{NeoTrace} and \textsc{KoAGE} let us collect Korean neologisms at scale and assess their acquisition and generation across 21 LLMs, several limitations remain, each of which we bound rather than remove.
First, a first-attestation date is an upper bound on emergence rather than its moment: a word may have circulated before its earliest retrieved use. Searching the top 200 relevance-ranked results per channel recovers the earliest sense-matched attestation within 1,000 results for 85\% of entries, against 53\% for a top-10 search, with a mean lag of half a year (Figure~\ref{fig:rankprec}b); a blind human check of 100 entries had found an earlier use for 16\%, the gap this search was introduced to close. For words of obscure origin, usage-based dating corrected the bounds inherited from lexicographic sources by 4.5 years on average, in line with corpus work that dates neologisms by their earliest observed use \citep{kerremans2012neocrawler,wuerschinger2021social}. On the model side, documented cutoffs may differ from effective ones \citep{cheng2024dated}; we check this through the post-cutoff-but-known rate but do not probe it directly.

Second, we rely on LLM judges at four points: admission of candidates, A1 scoring, G1 mechanism labeling, and G2 outcome labeling. Each was checked against a human rater on a blind sample, with agreement of 95.5\% on admission, 81.3\% ($\kappa$ 0.72; 89.3\% on exact vs.\ not) on A1, 80.3\% ($\kappa$ 0.74) on mechanisms, and perfect agreement on whether a G2 community formed a word (Appendix~\ref{app:judge}), which supports relying on them for the quantities we report. All results are single runs at vendor default temperature, since several current models accept no temperature argument; a temperature-0 re-run of four models moves aggregate accuracy by at most 3.6 points.

Third, \textsc{KoAGE} and its lexicon follow a taxonomy and task design made for Korean, but the collection method is not tied to the language. Only three components of \textsc{NeoTrace} are language-specific: the surface normalizer, the metalinguistic marker patterns, and the morphological analyzer; replacing them with English equivalents and running the framework on Reddit comments and online news accepted 210 and 160 neologisms, 98--100\% of them valid words and 87\% emerged after 2015 (Figure~\ref{fig:validation}b; Appendix~\ref{app:english}). A full benchmark for another language would still need its own taxonomy and reference coinages.

Finally, G2 shows that agent communities do coin words and that the frame decides whether they do, but with one backbone per community, one seed, and 25 cards whose reference coinages arose under particular social pressures. Because the social context and the neutral reminder both elicit and shape coinage, a stricter harness with controlled personas and more cards is needed before agents' coinage habits can be set against people's. Such a comparison matters beyond this benchmark: how agents coin and adopt words is part of how they form conventions within a community \citep{lewis2008convention,ashery2025emergent}, and so of whether they can take part in one as members rather than tools, which we leave to future work.

\section*{Ethics Statement}
The lexicon includes slang expressions, some of which may be pejorative. These entries are documented as linguistic data with source attribution and do not imply endorsement. The lexicon also includes a small number of terms related to sexual services and abuse, retained to support the recognition and moderation of potentially harmful content. Their definitions are strictly descriptive.
Data obtained through search APIs, YouTube, and text corpora were collected and used in accordance with the applicable terms of service and usage policies. The released code includes license information and documentation of the relevant policies. Released example sentences are drawn from publicly available news articles and monographs and contain no personal identifiers.
All simulated communities are fictional, and agents are explicitly instructed to engage in role-playing. Model APIs were used in accordance with their respective providers' policies.

\bibliography{custom}

\clearpage
\appendix

\section{\textsc{NeoTrace} details}
\label{app:collection}
\paragraph{Coverage beyond earlier collectors.} Unlike newspaper-only surveys and fixed-lexicon monitoring, \textsc{NeoTrace} combines an offline corpus, live search evidence and an evidence-grounded LLM verifier. Of 4,271 entries, 1,317 (31\%) are absent from the lexicographic references and come from the online or web tracks; 237 date to 2025--2026.

\paragraph{Shared resources.} The \emph{reference database} of known neologisms (11,514 distinct headwords) merges the five annual volumes of the \emph{Neologisms 2020--2024} monograph series (\citereports), two monographs on neologism formation \citep{choi2022neologism,nam2022neologism} and the ten NIKL annual neologism reports for 2008--2019 \citep{nikl2019collections}. It supplies the positives of the ranker, the marker and splinter patterns, and the deduplication reference; the NIKL reports are used only for seeds, temporal labeling and ranking, never as lexicon items.

The \emph{temporal exclusion lexicon} (TEL, 1.7M forms) combines static and dated layers. Two are static: the \emph{Standard Korean Language Dictionary} (2026-03 build, 437K headwords)\footnote{\url{https://stdict.korean.go.kr}, via \url{https://github.com/dahlia/ko-stdict-mcp}.} and Korean Wiktionary\footnote{\url{https://kaikki.org/dictionary/Korean/}.} are excluded at every date, except that a headword which is itself a reference neologism moves to the dated layer with its own year. Three are dated: high-frequency forms of the pre-2019 corpora are excluded from 2020 on, earlier neologisms only after their own year, and NIKL named entities (2021) and Wiktionary slang are soft-penalized rather than excluded.

The static layers can hide a word that the dictionary has codified since it emerged. The overlap is low in the reference sets, where only 10 of the 3,463 neologisms of 2020--2024 (0.3\%) and 109 of the 8,128 NIKL neologisms of 2008--2019 (1.3\%) appear in the 2026 dictionary, although false negatives among unobserved neologisms cannot be ruled out. Morphological features come from the Kiwi analyzer\footnote{\url{https://github.com/bab2min/Kiwi}}.

\begin{table}[h]
\centering\footnotesize
\setlength{\tabcolsep}{4pt}
\setlength{\tabcolsep}{3pt}
\fittable{\columnwidth}{%
\begin{tabular}{l r l}
    \toprule
    \textbf{Genre} & \textbf{Documents} & \textbf{Source} \\
    \midrule
    Newspapers (2019--2024) & 5,483,690 & Modu \\
    Video comments, captions (2013--2021) & 4,839,072 & AI Hub \\
    Messenger and SNS (2012--2021) & 1,973,763 & AI Hub \\
    Community posts, comments (2017--2024) & 495,622 & Modu \\
    Everyday spoken dialogue (2004--2025) & 43,872 & Modu \\
    \midrule
    Total & 12,836,019 & \\
    \bottomrule
\end{tabular}}
\caption{Online corpus by genre, with the production years of its documents. Modu = NIKL Modu corpora; AI Hub = Korean AI data hub. Every document keeps its production date and provenance (file, document id, URL); 9\% of the messenger documents carry no date and are used for usages only.}
\label{tab:corpus}
\end{table}

\paragraph{Online corpus: deterministic extraction.} The 12.8M-document corpus (Table~\ref{tab:corpus}), sourced from Modu corpus\footnote{\url{https://kli.korean.go.kr/corpus}} and AI Hub\footnote{\url{https://aihub.or.kr}}, is searched with three extractors: (i) 44 productive splinters induced from reference neologisms (e.g., \gl{-린이}{novice} and \gl{-세권}{within reach of}), including newly discovered splinters; (ii) nine regular-expression metalinguistic markers over quoted forms (e.g., \gl{`X'라는 신조어}{the neologism X} and \gl{줄여서 `X'}{X for short}); and (iii) out-of-vocabulary forms after stripping particles/endings, plus previously unseen known-word bigrams with high normalized PMI. Marker precision against the reference varies (4.2--72.7\%), so it is evidence for ranking rather than an admission rule. Kiwi analysis provides additional morphological evidence. These extractors yield 3.9M candidates before filtering, including names and typos.

\begin{table}[h]
\centering\footnotesize
\setlength{\tabcolsep}{4pt}
\fittable{\columnwidth}{%
\begin{tabular}{l p{0.70\columnwidth}}
    \toprule
    \textbf{Group} & \textbf{Features} \\
    \midrule
    Distribution (6) & log frequency, log document count, number of sources, share of occurrences from 2021 on, share of undated occurrences, first year \\
    \midrule
    Surface form (5) & length, Latin-script mixing, digit mixing, number of distinct particle attachments, soft named-entity score \\
    \midrule
    Extraction evidence (7) & marker score, marker document count, number of marker kinds, number of splinter families matched, MWE flag, MWE normalized PMI, number of extractors capturing the candidate \\
    \bottomrule
\end{tabular}}
\caption{The 18 ranking features; four Kiwi morphological classes are added in the second training stage.}
\label{tab:features}
\end{table}

\paragraph{Online corpus: ranking.} Table~\ref{tab:features} lists the 18 features. The number of extractors that captured a candidate is the strongest: among the 1,987 judged candidates, 5\% of those captured by one extractor are neologisms, 24\% by two and 76\% by three. A positive--unlabeled gradient-boosting ranker is trained in two stages (all candidates on the statistical features, then the top 200K with the four morphological classes added) with the reference database as positives; as a diagnostic of ranking quality, 78\% of held-out reference items fall in the top 100K (2.6\% of 3.9M candidates); the judged pool of 2,000 is chosen for precision. Precision down the ranking (Figure~\ref{fig:rankprec}) was measured by verifying every candidate in the top 2,000 and, beyond that, a stratified sample of 100 per rank window with the same two-round pipeline.

\begin{table*}[t!]
\centering
\footnotesize
\fittable{\textwidth}{%
\begin{tabular}{l >{\raggedright\arraybackslash}p{4.1cm} >{\raggedright\arraybackslash}p{4.5cm} >{\raggedright\arraybackslash}p{3.6cm}}
    \toprule
    & \textbf{Korean (this paper)} & \textbf{English, Reddit} & \textbf{English, news} \\
    \midrule
    \addlinespace[2pt]
    \rowcolor{gray!15}\multicolumn{4}{l}{\textit{Components (unchanged unless noted)}} \\
    Normalizer & particle and ending stripping & lower-casing, inflection merge & same \\
    \addlinespace[2pt]
    Exclusion lexicon & dictionary; pre-2019 corpus forms; dated past neologisms & pre-2019 Reddit forms ($\geq$20); NEO-BENCH, dated; Wiktionary as a feature & pre-2019 news forms; otherwise same \\
    \addlinespace[2pt]
    Marker patterns & 9 Korean regexes (\ko{`X'라는 신조어}, \ldots) & 9 English regexes (\emph{the term ``X''}, \emph{so-called X}, \emph{X is slang for}, \ldots) & same \\
    \addlinespace[2pt]
    Novelty feature & Kiwi analysis class & lemmatization failure, sub-word count & same \\
    \addlinespace[2pt]
    Ranker, judge, dating & PU ranker; two-round LLM judge on evidence packets; observed minimum & same code & same code \\
    \midrule
    \rowcolor{gray!15}\multicolumn{4}{l}{\textit{Funnel}} \\
    \addlinespace[2pt]
    Documents / tokens & 12.8M documents & 152M comments / 3.9B & 22.5M articles / 5.7B \\
    \addlinespace[2pt]
    Candidates ($\geq$5 occ.) & 3.9M & 1.12M & 1.82M \\
    \addlinespace[2pt]
    Adjudicated (top-$K$) & 2,000 & 2,000 & 2,000 \\
    \addlinespace[2pt]
    Accepted after verification & 386 (239 kept) & 210 & 160 \\
    \addlinespace[2pt]
    Precision, re-verified & 95.5\% (human, 200 stratified) & 98\% real, 87\% post-2015 (web, 100) & 100\% real, 87\% post-2015 (web, 100) \\
    \addlinespace[2pt]
    Reference recall & 78\% @100K (held-out) & 42\% @20K, 60\% @50K (551 present) & 43\% @20K, 55\% @50K (495 present) \\
    \addlinespace[2pt]
    Accepted, by type & blends and abbreviations 59\% of single words & derivation 26, blend 20, coinage 19, compound 17, abbreviation 10\% & blend 35, compound 23, derivation 17, coinage 12\% \\
    \bottomrule
\end{tabular}}
\caption{\textsc{NeoTrace} on English. Reference recall counts NEO-BENCH items occurring at least five times in the corpus; ``post-2015'' counts re-verified items whose current sense emerged in 2015 or later. Korean column: the online-corpus track of Section~\ref{sec:neotrace}.}
\label{tab:english}
\end{table*}

\paragraph{Online corpus: verification.} The evidence packet holds usages stratified by year and register, TEL status of the word and its components, the yearly frequency series and NAVER blog\footnote{\url{https://section.blog.naver.com/}} and Q\&A\footnote{\url{https://kin.naver.com/}} results (news search is excluded as circular). The judge returns a status (neologism, existing, proper name, typo or noise, nonce, institutional term, uncertain), a definition, flags (institutional, onomastic origin, embedded neologism), cited evidence ids and a confidence. Decision rules treat official names given by institutions as non-neologisms but folk nicknames and abbreviations of them as neologisms (\gl{노란봉투법}{yellow-envelope act}, \ko{검수완박}, an initialism), count borrowings that entered Korean discourse around the first year, class short-lived coinages as ephemeral neologisms and one-speaker jokes as nonce, and reject a candidate whose earliest dated web occurrence precedes the corpus first year by two or more years.

A second round re-judges every decisive verdict with confidence below 0.75 and a random 10\% of the confident ones (589 of the 2,000) with a different vendor's model; agreements are accepted, disagreements adjudicated, double-uncertain items set aside as ambiguous. Round-to-round reproducibility is 91.4\% for confident decisions and 80.6\% overall. The reported run accepted 386 neologisms, 1,511 hard negatives, 82 institutional terms and 8 ambiguous items from the top 2,000; 13 candidates received no final label and enter neither the lexicon nor A2 distractors. After merging tracks, 239 of the 386 accepted candidates remain; 1,735 rejected items across tracks supply A2 distractors.

\paragraph{Web API.} 290 query combinations over NAVER news\footnote{\url{https://news.naver.com/}}, blog and Q\&A search (2019--2026; 60 topic queries such as \gl{요즘 신조어}{recent neologisms}, plus the marker patterns as queries) return 207,457 snippets, and the YouTube Data API\footnote{\url{https://developers.google.com/youtube/v3}.} supplies titles, descriptions and captions of 760 popular videos from 21 channels. The marker and OOV extractors run on the snippets; verification is the same two-round adjudication, with the web evidence in the packet, and the first observation is recorded as an observed minimum when no dated source is found.

\paragraph{Admission.} All tracks pass the same admission: cross-source deduplication, taxonomy labeling, metadata (stratum, formation mechanism, transparency, per-model knowledge probe) and a dated first attestation; entries without a year are dropped. Items labeled \emph{report} come from the \emph{Neologisms 2020--2024} series and the two monographs, with definitions written by lexicographers; web-collected items come from forum, SNS and video text. Membership of the lexicon is decided by the pipeline alone.

\paragraph{Dating.} Each entry carries a first-attestation year and its evidential basis, categorized as \emph{recorded}, \emph{observed-minimum}, or \emph{web}. For \emph{recorded} entries, the year is obtained from the report or monograph listing the word or from a dated corpus document. For \emph{observed-minimum} entries, it is determined by the earliest sense-matched occurrence in our corpora. For \emph{web} entries, it is based on the earliest sense-matched attestation identified among top-200 relevance-ranked NAVER news and blog search results. To validate the observed-minimum dates, we compare the earliest corpus occurrences with expert-assigned dates for report items that also appear in our corpora. This comparison assesses the agreement between corpus-based and expert-assigned dates.

\paragraph{Search depth.} The depth of 200 was set on a stratified sample of 200 entries searched to 1,000 results per channel in both relevance and newest-first order, with every candidate snippet judged for sense match against the definition. In relevance order the 1,000-result minimum is reached for 53\% of entries at $k$=10, 79\% at 100, 85\% at 200 and 88\% at 300, with a mean lag of 0.55 years at 200, and gains fall below 5 points per further 100 results beyond 200 (Figure~\ref{fig:rankprec}b); adding the newest-first order finds no earlier attestations, so only relevance order is used.

\subsection{Transfer to English}
\label{app:english}

The framework is Korean-specific in three components: the surface normalizer (particle and ending stripping), the metalinguistic marker patterns and the morphological analyzer used as a novelty feature. To check that the rest transfers, we re-ran the online-corpus track on English with those components replaced and everything else unchanged: the layered temporal exclusion lexicon, the four deterministic candidate generators, the evidence vector and positive--unlabeled ranker, the evidence-packet adjudication by an LLM with a second-vendor audit, and observed-minimum dating.

Two genres were used, matching the community register of our web API track and the news register of the \emph{Neologisms 2020--2024} series: Reddit\footnote{\url{https://www.reddit.com/}} comments (the first 200\,MB of each monthly Pushshift dump, 2019--2025) and CC-News English articles\footnote{\url{https://huggingface.co/datasets/stanford-oval/ccnews}} (one monthly shard, 2019--2024). The temporal baseline is the same corpus two to three years earlier (2016--2018 Reddit; 2017--2018 news), and the reference list of neologisms is NEO-BENCH \citep{zheng2024neobench}. Table~\ref{tab:english} gives collection and validation results.

The English run accepts 210 Reddit and 160 news forms, with 98--100\% confirmed as real words and 87\% first attested since 2015 in 100-item web checks per genre (Table~\ref{tab:english}). Accepted types differ by genre: Reddit favors slang and meme formations, while news favors blends and compounds. Reference recall is lower than Korean (42--43\% vs.\ 78\%): NEO-BENCH includes many multiword expressions, but the English extractor targets only new two-word sequences, yielding 12--14\% recall for that group versus 40--76\% for single words. Since the run samples Reddit/news data and omits the web API track, it demonstrates component portability rather than a complete English lexicon.

\begin{table*}[t!]
\centering\small
\setlength{\tabcolsep}{4pt}
\fittable{\textwidth}{%
\begin{tabular}{l l p{4.6cm} p{5.6cm}}
    \toprule
    \textbf{Class} & \textbf{Leaf} & \textbf{Criterion} & \textbf{Examples (first attestation)} \\
    \midrule
    \multirow{5}{*}{FORM} & CPD compounding & root + root, both intact & \gl{폭염거지}{heat-wave pauper} (2026) \\
    \addlinespace[2pt]
     & DER derivation & affix attached to a base & \gl{하이볼족}{highball tribe} (2024) \\
     \addlinespace[2pt]
     & BLD blending & fragments of two words combined (incl.\ productive splinters such as \gl{-린이}{novice}, \gl{-세권}{within reach of}) & \ko{맵파민} \emph{spicy}+\emph{dopamine} (2024) \\
     \addlinespace[2pt]
     & ABB abbreviation & initial syllables of a phrase, or clipping & \gl{마용성길}{Mayongseong Street} $<$ \ko{마포 용산 성동 길음} (2026) \\
     \addlinespace[2pt]
     & SYN syntactic fusion & particle or ending frozen inside the word & \gl{없던아빠}{the dad there never was} (2023) \\
     \addlinespace[2pt]
    \midrule
    \multirow{3}{*}{MWE} & NSEQ noun sequence & juxtaposed nouns forming a fixed unit & \gl{혁신도시 기러기}{innovation-city goose dad} (2026) \\
    \addlinespace[2pt]
     & MOD modifier & adnominal + noun & \gl{딸깍 작품}{one-click artwork} (2026) \\
     \addlinespace[2pt]
     & CLS clausal & frozen clause or sentence & \gl{유금무죄 무금유죄}{badge acquits, no badge convicts} (2024) \\
     \addlinespace[2pt]
    \midrule
    BORROW & -- & imported whole from another language, incl.\ transliteration and acronyms & \ko{롱제비티} \emph{longevity} (2026) \\
    \addlinespace[2pt]
    \midrule
    COIN & -- & no pre-existing material: sound or script play & \gl{삐끼삐끼}{Pikki Pikki} (2024) \\
    \bottomrule
\end{tabular}}
\caption{Taxonomy leaves with criteria and attested examples from the lexicon.}
\label{tab:taxonomy}
\end{table*}

\begin{table*}[t!]
\centering
\resizebox{\textwidth}{!}{%
\begin{tabular}{l rrrrr rrr r r r}
    \toprule
    & \multicolumn{5}{c}{\textbf{FORM} (Morphological)} & \multicolumn{3}{c}{\textbf{MWE} (Multiword)}
    & \multirow{2}{*}{\textbf{BORROW}} & \multirow{2}{*}{\textbf{COIN}} & \multirow{2}{*}{\textbf{Total}} \\
    \cmidrule(lr){2-6} \cmidrule(lr){7-9}
    \textbf{Source} & CPD & DER & BLD & ABB & SYN & NSEQ & MOD & CLS & & & \\
    \midrule
    Literature             & 285 & 355 & 562 & 435 & 20 & 1,078 & 71 & 14 & 120 & 14 & 2,954 \\
    Online corpus          & 54 & 37 & 64 & 37 & 7 & -- & 1 & 1 & 34 & 4 & 239 \\
    Web API                & 158 & 97 & 244 & 141 & 10 & 234 & 29 & 28 & 126 & 11 & 1,078 \\
    \midrule
    \textbf{Total}         & \textbf{497} & \textbf{489} & \textbf{870} & \textbf{613} & \textbf{37} & \textbf{1,312} & \textbf{101} & \textbf{43} & \textbf{280} & \textbf{29} & \textbf{4,271} \\
    \bottomrule
\end{tabular}}
\caption{Composition of the \textsc{KoAGE} lexicon (4,271 entries) by collection track and taxonomy leaf.
CPD: compounding; DER: derivation; BLD: blending (incl.\ splinters);
ABB: abbreviation (initial-syllable and clipping); SYN: syntactic fusion;
NSEQ: noun sequence; MOD: modifier construction; CLS: clausal expression.
``--'' denotes zero.}
\label{tab:lexicon-composition}
\end{table*}

\subsection{Taxonomy of neologism types}
\label{app:taxonomy}

\paragraph{Design.} The four top-level classes distinguish where the newness arises; their ten mutually exclusive leaves distinguish the formation or syntactic mechanism (Table~\ref{tab:taxonomy}). FORM follows Korean morphological classifications \citep{choi2022neologism}; MWE handles fixed phrases, while BORROW and COIN distinguish whole-word imports and non-compositional creations \citep{nam2022neologism,algeo1991fifty}. For boundary cases, material combined by Korean speakers is FORM (\gl{호캉스}{hotel+vacance}), but a word imported whole is BORROW (\gl{워케이션}{workation}); fused single words are FORM-SYN, whereas independently written phrases are MWE-CLS.

\subsection{Lexicon composition}
\label{app:composition}

Table~\ref{tab:lexicon-composition} cross-tabulates the four classes and ten taxonomy leaves by collection source. Each of the 4,271 entries occupies one taxonomy leaf and one source row; source identifies its collection route, not its linguistic type. The lexicon is dominated by FORM (2,506 entries, 59\%) and MWE (1,456, 34\%). Blending (870) and abbreviation (613) together make up 59\% of FORM, while noun sequences account for 1,312 of the MWE entries. BORROW (280) and COIN (29) are small.

The source-by-type breakdown further shows the complementarity of the three collection tracks. The literature track contributes 2,954 entries, concentrated in FORM and MWE; the online-corpus track contributes 239, and the web API track adds 1,078, including 244 blends and 141 abbreviations. Collection provenance and taxonomy are therefore separate attributes, rather than a one-to-one mapping between a track and a linguistic phenomenon.

\begin{table*}[t!]
\centering\scriptsize
\setlength{\tabcolsep}{4pt}
\resizebox{\textwidth}{!}{%
\begin{tabular}{l l c c l c l l}
    \toprule
    \textbf{Name in tables} & \textbf{API identifier / HF repository} & \textbf{Release} & \textbf{Cutoff} & \textbf{Decoding} & \textbf{Weights} & \textbf{A1 judge} & \textbf{Ref.} \\
    \midrule
    \rowcolor{gray!15}\multicolumn{8}{l}{\textit{Proprietary (API)}} \\
    GPT-3.5 Turbo & \texttt{gpt-3.5-turbo-0125} & 2023-03 & 2021-09 & vendor default $T$ & API & Claude Opus 5 (low) & \citep{ouyang2022instructgpt} \\
    GPT-4o & \texttt{gpt-4o-2024-05-13} & 2024-05 & 2023-10 & vendor default $T$ & API & Claude Opus 5 (low) & \citep{openai2024gpt4o} \\
    GPT-5 & \texttt{gpt-5-2025-08-07} & 2025-08 & 2024-09 & vendor default $T$, effort low & API & Claude Opus 5 (low) & \citep{openai2025gpt5} \\
    GPT-5.4 mini & \texttt{gpt-5.4-mini-2026-03-17} & 2026-03 & 2025-08 & vendor default $T$, effort low & API & Claude Opus 5 (low) & \citep{openai2026gpt54} \\
    GPT-5.6 sol & \texttt{gpt-5.6-sol} & 2026-07 & 2026-02 & vendor default $T$, effort low & API & Claude Opus 5 (low) & \citep{openai2026gpt56} \\
    Claude Sonnet 4.5 & \texttt{claude-sonnet-4-5-20250929} & 2025-09 & 2025-07 & vendor default $T$, thinking off & API & GPT-5.6 luna & \citep{anthropic2025sonnet45} \\
    Claude Sonnet 5 & \texttt{claude-sonnet-5} & 2026-06 & 2026-01 & vendor default $T$, effort low & API & GPT-5.6 luna & \citep{anthropic2026sonnet5} \\
    Gemini 3 Flash & \texttt{gemini-3-flash-preview} & 2025-12 & 2025-01 & vendor default $T$, thinking low & API & Claude Opus 5 (low) & \citep{google2025gemini3} \\
    Gemini 3.5 Flash-Lite & \texttt{gemini-3.5-flash-lite} & 2026-07 & 2026-03 & vendor default $T$, thinking low & API & Claude Opus 5 (low) & \citep{google2026gemini35} \\
    Solar Pro 2 & \texttt{solar-pro2-251215} & 2025-12 & 2025-03 & vendor default $T$, reasoning low & API & Claude Opus 5 (low) & \citep{upstage2026solar} \\
    Solar Pro 4 & \texttt{solar-pro4-260806} & 2026-08 & 2026-02 & vendor default $T$, reasoning low & API & Claude Opus 5 (low) & \citep{upstage2026solar} \\
    HCX-003 & \texttt{HCX-003} & 2024-01 & -- & vendor default $T$ & API & Claude Opus 5 (low) & \citep{naver2024hyperclovax} \\
    HCX-007 & \texttt{HCX-007} & 2025-07 & -- & vendor default $T$, thinking low & API & Claude Opus 5 (low) & \citep{naver2024hyperclovax} \\
    \rowcolor{gray!15}\multicolumn{8}{l}{\textit{Open-weight}} \\
    Qwen2 7B & \texttt{Qwen/Qwen2-7B-Instruct} & 2024-06 & -- & greedy & bf16 & Claude Opus 5 (low) & \citep{yang2024qwen2} \\
    Qwen2.5 7B & \texttt{Qwen/Qwen2.5-7B-Instruct} & 2024-09 & -- & greedy & bf16 & Claude Opus 5 (low) & \citep{yang2024qwen25} \\
    Qwen3 8B & \texttt{Qwen/Qwen3-8B} & 2025-04 & -- & greedy, thinking off & bf16 & Claude Opus 5 (low) & \citep{yang2025qwen3} \\
    Qwen3.5 9B & \texttt{Qwen/Qwen3.5-9B} & 2026-02 & -- & greedy, thinking off & bf16 & Claude Opus 5 (low) & \citep{qwen2026qwen35} \\
    EXAONE 3.0 7.8B & \texttt{LGAI-EXAONE/EXAONE-3.0-7.8B-Instruct} & 2024-08 & -- & greedy & bf16 & Claude Opus 5 (low) & \citep{lgai2024exaone3} \\
    EXAONE 3.5 7.8B & \texttt{LGAI-EXAONE/EXAONE-3.5-7.8B-Instruct} & 2024-12 & -- & greedy & bf16 & Claude Opus 5 (low) & \citep{lgai2024exaone35} \\
    EXAONE 4.0 32B & \texttt{LGAI-EXAONE/EXAONE-4.0-32B} & 2025-07 & -- & greedy, thinking off & NF4 4-bit & Claude Opus 5 (low) & \citep{lgai2025exaone4} \\
    EXAONE 4.5 33B & \texttt{LGAI-EXAONE/EXAONE-4.5-33B} & 2026-04 & 2024-12 & greedy, thinking off & NF4 4-bit & Claude Opus 5 (low) & \citep{lgai2026exaone45} \\
    \bottomrule
\end{tabular}}
\caption{Evaluated models: API identifier or Hugging Face repository, release and cutoff months (``--'' = unpublished; release month used), decoding, weights, A1 judge and reference.}
\label{tab:models}
\end{table*}

\begin{table}[t!]
\centering\footnotesize
\resizebox{\columnwidth}{!}{%
\begin{tabular}{l r >{\raggedright\arraybackslash}p{0.56\columnwidth}}
\toprule
\textbf{Task} & \textbf{Items} & \textbf{Admitted items} \\
\midrule
A1-W & 2,504 & evaluation pool: entries first attested 2020 or later \\
A1-WU & 1,819 & pool entries with a reviewed usage sentence \\
A2 & 1,230 & usage masks without leaking the answer; all three distractors pass the adversarial check \\
A3 & 1,332 & FORM entries plus controls from BORROW and COIN \\
G1 & 839 & pool entries whose word is not a plain concatenation of its materials \\
G2 & 25 & cards with a verified origin (community, medium, situation) \\
\bottomrule
\end{tabular}}
\caption{Item counts per task and the admission rules.}
\label{tab:taskcounts}
\vspace{-11pt}
\end{table}

\section{Models and settings}
\label{app:settings}

Table~\ref{tab:models} lists every model with the exact identifier used, so that names such as ``GPT-5.6 sol'' or ``Gemini 3 Flash'' (served only as \texttt{gemini-3-flash-preview}) can be mapped to a specific snapshot. GPT-5.6 sol, Claude Sonnet 5, the two Gemini models and the two HyperCLOVA X models return no dated snapshot. Anthropic and Google have retired their pre-2025 models, so within-family comparison over time is possible only for OpenAI and Upstage.

\begin{table*}[t!]
\centering\small
\setlength{\tabcolsep}{3pt}
\resizebox{\textwidth}{!}{%
\begin{tabular}{l l r r r}
    \toprule
    \textbf{Judgment} & \textbf{Sample} & $n$ & \textbf{Agree} & $\kappa$ \\
    \midrule
    \rowcolor{gray!15}\multicolumn{5}{l}{\textit{A1 definition match (0/1/2)}} \\
    Judge vs.\ reference & known-answer pairs (156 approved + 410 controls) & 566 & 96--97 & 0.91--0.93 (bin.) \\
    Judge vs.\ judge (Opus 5 vs.\ GPT-5.6) & stratified Sonnet 5 outputs & 300 & 79--81 & 0.68--0.72 \\
    Judge vs.\ human & random, stratified by judge score, all models & 150 & 81.3 & 0.72 (lin.\ 0.79) \\
    \quad binary (exact vs.\ not) & same & 150 & 89.3 & 0.77 \\
    \rowcolor{gray!15}\multicolumn{5}{l}{\textit{G1 mechanism label (four of the five classes)}} \\
    Classifier (Opus 5) vs.\ human & 75 per mechanism, four mechanisms & 300 & 80.3 & 0.74 \\
    Opus 5 vs.\ GPT-5.6 sol & same & 300 & 89.0 & 0.85 \\
    \rowcolor{gray!15}\multicolumn{5}{l}{\textit{G2 exit-interview judgment}} \\
    Formation (name shared by $\geq$2 agents) & stratified sessions & 60 & 100 & 1.00 \\
    Consensus count (0--4, exact) & same & 60 & 93 & 0.88 (lin.\ 0.95) \\
    Reproduction (exact or near) & same & 60 & 98 & 0.88 \\
    Relation, near boundary & Sonnet 5 ablations: all judged \emph{near} & 27 & 85 & n/a \\
    \bottomrule
    \end{tabular}}
\caption{Judge validation against one human rater: agreement (\%) and Cohen's $\kappa$ (lin.\ = linearly weighted, bin.\ = exact vs.\ not). The near-boundary row checks one judge label, so $\kappa$ is undefined.}
\label{tab:judgeval}
\end{table*}

\paragraph{Task construction.} Table~\ref{tab:taskcounts} gives the item counts and the rule that admits an item to each task. All tasks start from the evaluation pool, the lexicon entries first attested in 2020 or later; the usage-based tasks are restricted to entries with a reviewed usage sentence, and A2 further requires that the sentence can be masked without exposing a fragment of the answer and that every distractor passes an adversarial check by a second-vendor model, which verifies that no distractor fits the blank; a flagged distractor is replaced by another of the same kind (same-formation distractors are drawn from entries of the same taxonomy leaf and similar length). Abbreviations are the main casualty of the masking rule, because their source phrase often appears in the same sentence. The A3 answer is the entry's taxonomy leaf, assigned at admission and revised in the taxonomy audit. A3-WD definitions are rewritten to remove phrasing that reveals the mechanism (62 of 1,332 still hint at it; excluding them changes no mean by more than 0.8 points).

\paragraph{Prompts and batching.} Every acquisition task and G1 use one fixed Korean instruction per task, identical for all models (Appendix~\ref{app:prompts}), with ten items per request and a JSON-array output format. Requests whose JSON cannot be parsed are retried twice and re-issued item by item; items that return nothing are scored as unanswered, never silently dropped. Prompts are sent as a single user turn; no system prompt is used. The A1 instruction permits a ``don't know'' answer, which is scored as incorrect.

\paragraph{Decoding.} Proprietary models are called at the vendor default temperature, because several current models accept no temperature argument. The four that do accept one (GPT-4o, Claude Sonnet 4.5, Gemini 3 Flash, Solar Pro 2) were re-run at temperature 0 on the first 500 A1-W and A1-WU items with the same judge, which moved aggregate accuracy by at most 3.6 points in either direction while item-level agreement between the two runs was 79--88\%. Reasoning models run at their lowest reasoning setting: \texttt{reasoning\_effort=low} (OpenAI, Upstage), \texttt{effort=low} (Claude Sonnet 5), \texttt{thinking\_level=low} (Gemini 3.x), thinking on with low effort (HCX-007); Claude Sonnet 4.5 runs without extended thinking. Output caps are 4,000 tokens per ten-item request (8,000 for Upstage reasoning models) and 1,200 tokens per G2 turn (4,000 for Upstage). Open-weight models decode greedily with the model's own chat template, with thinking disabled for Qwen3/3.5 and EXAONE 4.x.

\paragraph{Judging.} A1 is judged by Claude Opus 5 at low effort through the Message Batches API (GPT-5.6 luna for Claude-family models, to avoid same-vendor judging), G1 mechanism labels by Claude Opus 5 at low effort, and G2 exit interviews by GPT-5.6 sol; validation is in Appendix~\ref{app:judge}. Judges see 25 items per request and return JSON.

\paragraph{Open-weight models.} All local runs use one NVIDIA RTX A6000 (48\,GB) per model with PyTorch 2.13 and \texttt{transformers} 5.16; EXAONE 3.0/3.5, whose code predates \texttt{transformers}~5, run with PyTorch 2.6 and \texttt{transformers} 4.46. 7--9B models are loaded in bfloat16, the 32B/33B EXAONE models in 4-bit NF4 with bfloat16 compute, and EXAONE 4.5 through its vision--language class with the text path only. Generation is batched with left padding and per-model batch sizes. Scripts and the per-model batch table are released with the code.

\begin{table*}[t!]
\centering\small
\setlength{\tabcolsep}{4pt}
\resizebox{\textwidth}{!}{%
\begin{tabular}{l cc cc cc cc cc c c}
\toprule
& \multicolumn{2}{c}{A1-W} & \multicolumn{2}{c}{A1-WU} & \multicolumn{2}{c}{A2} & \multicolumn{2}{c}{A3-W} & \multicolumn{2}{c}{A3-WD} & \multicolumn{2}{c}{G2} \\
\cmidrule(lr){2-3} \cmidrule(lr){4-5} \cmidrule(lr){6-7} \cmidrule(lr){8-9} \cmidrule(lr){10-11} \cmidrule(lr){12-13}
\textbf{Model} & pre & post & pre & post & pre & post & pre & post & pre & post & cards & Formed CI \\
\midrule
GPT-3.5 Turbo & 931 ($\pm$1) & 1,474 ($\pm$0) & 703 ($\pm$2) & 1,075 ($\pm$1) & 484 ($\pm$4) & 713 ($\pm$4) & 415 ($\pm$4) & 838 ($\pm$3) & 415 ($\pm$4) & 838 ($\pm$3) & 25 & 0--13 \\
GPT-4o & 1,832 ($\pm$2) & 590 ($\pm$2) & 1,383 ($\pm$3) & 411 ($\pm$4) & 969 ($\pm$2) & 244 ($\pm$5) & 913 ($\pm$3) & 360 ($\pm$5) & 913 ($\pm$3) & 360 ($\pm$5) & 25 & 9--39 \\
GPT-5 & 2,174 ($\pm$2) & 241 ($\pm$5) & 1,641 ($\pm$2) & 164 ($\pm$8) & 1,159 ($\pm$1) & 64 ($\pm$7) & 1,107 ($\pm$3) & 156 ($\pm$8) & 1,107 ($\pm$3) & 156 ($\pm$6) & 22 & 67--95 \\
GPT-5.4 mini & 2,335 ($\pm$2) & 149 ($\pm$6) & 1,703 ($\pm$2) & 100 ($\pm$9) & 1,184 ($\pm$2) & 41 ($\pm$12) & 1,221 ($\pm$3) & 99 ($\pm$10) & 1,221 ($\pm$3) & 99 ($\pm$9) & 23 & 45--81 \\
GPT-5.6 sol & 2,422 ($\pm$2) & 82 ($\pm$8) & 1,766 ($\pm$2) & 53 ($\pm$13) & 1,208 ($\pm$1) & 22 ($\pm$7) & 1,280 ($\pm$2) & 52 ($\pm$12) & 1,280 ($\pm$2) & 52 ($\pm$10) & 19 & 83--100 \\
Claude Sonnet 4.5 & 2,327 ($\pm$2) & 157 ($\pm$5) & 1,698 ($\pm$2) & 105 ($\pm$9) & 1,182 ($\pm$2) & 43 ($\pm$10) & 1,216 ($\pm$3) & 104 ($\pm$9) & 1,216 ($\pm$3) & 104 ($\pm$9) & 21 & 77--99 \\
Claude Sonnet 5 & 2,415 ($\pm$2) & 89 ($\pm$7) & 1,764 ($\pm$2) & 55 ($\pm$13) & 1,208 ($\pm$1) & 22 ($\pm$14) & 1,277 ($\pm$3) & 55 ($\pm$13) & 1,277 ($\pm$2) & 55 ($\pm$11) & 24 & 80--99 \\
Gemini 3 Flash & 2,275 ($\pm$2) & 209 ($\pm$6) & 1,664 ($\pm$2) & 139 ($\pm$8) & 1,169 ($\pm$1) & 56 ($\pm$6) & 1,181 ($\pm$2) & 139 ($\pm$8) & 1,181 ($\pm$2) & 139 ($\pm$7) & 18 & 82--100 \\
Gemini 3.5 Flash-Lite & 2,432 ($\pm$2) & 72 ($\pm$8) & 1,773 ($\pm$2) & 46 ($\pm$11) & 1,209 ($\pm$1) & 21 ($\pm$13) & 1,287 ($\pm$3) & 45 ($\pm$14) & 1,287 ($\pm$3) & 45 ($\pm$13) & 24 & 74--98 \\
Solar Pro 2 & 2,289 ($\pm$2) & 195 ($\pm$2) & 1,673 ($\pm$2) & 130 ($\pm$8) & 1,173 ($\pm$2) & 52 ($\pm$11) & 1,193 ($\pm$3) & 127 ($\pm$8) & 1,193 ($\pm$3) & 127 ($\pm$9) & 24 & 47--82 \\
Solar Pro 4 & 2,422 ($\pm$2) & 82 ($\pm$3) & 1,766 ($\pm$2) & 53 ($\pm$12) & 1,208 ($\pm$1) & 22 ($\pm$14) & 1,280 ($\pm$3) & 52 ($\pm$13) & 1,280 ($\pm$3) & 52 ($\pm$12) & 24 & 60--91 \\
HCX-003 & 2,010 ($\pm$2) & 405 ($\pm$3) & 1,491 ($\pm$3) & 314 ($\pm$5) & 1,046 ($\pm$2) & 177 ($\pm$6) & 1,015 ($\pm$3) & 248 ($\pm$5) & 1,015 ($\pm$3) & 248 ($\pm$5) & 24 & 15--49 \\
HCX-007 & 2,327 ($\pm$2) & 157 ($\pm$5) & 1,698 ($\pm$2) & 105 ($\pm$8) & 1,182 ($\pm$2) & 43 ($\pm$14) & 1,216 ($\pm$3) & 104 ($\pm$8) & 1,216 ($\pm$3) & 104 ($\pm$9) & 24 & 47--82 \\
Qwen2 7B & 2,171 ($\pm$1) & 244 ($\pm$1) & 1,639 ($\pm$2) & 166 ($\pm$4) & 1,158 ($\pm$3) & 65 ($\pm$12) & 1,106 ($\pm$2) & 157 ($\pm$7) & 1,106 ($\pm$2) & 157 ($\pm$7) & 25 & 4--30 \\
Qwen2.5 7B & 2,174 ($\pm$1) & 241 ($\pm$1) & 1,641 ($\pm$2) & 164 ($\pm$6) & 1,159 ($\pm$3) & 64 ($\pm$12) & 1,107 ($\pm$3) & 156 ($\pm$7) & 1,107 ($\pm$2) & 156 ($\pm$7) & 25 & 0--13 \\
Qwen3 8B & 2,294 ($\pm$1) & 190 ($\pm$1) & 1,676 ($\pm$2) & 127 ($\pm$6) & 1,174 ($\pm$2) & 51 ($\pm$13) & 1,195 ($\pm$3) & 125 ($\pm$8) & 1,195 ($\pm$3) & 125 ($\pm$8) & 25 & 17--52 \\
Qwen3.5 9B & 2,422 ($\pm$1) & 82 ($\pm$2) & 1,766 ($\pm$2) & 53 ($\pm$8) & 1,208 ($\pm$2) & 22 ($\pm$16) & 1,280 ($\pm$2) & 52 ($\pm$12) & 1,280 ($\pm$3) & 52 ($\pm$13) & 25 & 27--63 \\
EXAONE 3.0 7.8B & 2,173 ($\pm$2) & 242 ($\pm$2) & 1,641 ($\pm$2) & 164 ($\pm$6) & 1,159 ($\pm$3) & 64 ($\pm$12) & 1,106 ($\pm$2) & 157 ($\pm$7) & 1,106 ($\pm$2) & 157 ($\pm$7) & 25 & 6--35 \\
EXAONE 3.5 7.8B & 2,180 ($\pm$2) & 235 ($\pm$2) & 1,646 ($\pm$2) & 159 ($\pm$6) & 1,160 ($\pm$2) & 63 ($\pm$12) & 1,110 ($\pm$2) & 153 ($\pm$7) & 1,110 ($\pm$2) & 153 ($\pm$7) & 25 & 41--77 \\
EXAONE 4.0 32B & 2,327 ($\pm$0) & 157 ($\pm$1) & 1,698 ($\pm$2) & 105 ($\pm$6) & 1,182 ($\pm$2) & 43 ($\pm$14) & 1,216 ($\pm$2) & 104 ($\pm$8) & 1,216 ($\pm$2) & 104 ($\pm$9) & 25 & 1--20 \\
EXAONE 4.5 33B & 2,180 ($\pm$2) & 235 ($\pm$2) & 1,646 ($\pm$2) & 159 ($\pm$6) & 1,160 ($\pm$2) & 63 ($\pm$12) & 1,110 ($\pm$3) & 153 ($\pm$7) & 1,110 ($\pm$3) & 153 ($\pm$7) & 25 & 11--43 \\
\bottomrule
\end{tabular}}
\caption{Item-level 95\% confidence intervals behind Table~\ref{tab:main}: for each acquisition cell, the number of items first attested before (pre) or after (post) the model's cutoff (boundary-year items without a month excluded) and, in parentheses, the half-width of the 95\% Wilson interval of its accuracy in points. The G2 columns give the cards unknown to the backbone and the card-level 95\% Wilson interval of the formation rate.}
\label{tab:seen}
\end{table*}

\section{Judge validation}
\label{app:judge}

\paragraph{Annotators.} Human annotation was carried out by three native Korean speakers, including a Korean-language expert, paid at the statutory minimum wage\footnote{The minimum wage in South Korea for 2026 is KRW 10,320 per hour.}. The blind validation of collected items and of judge outputs was performed by the expert.
Following the reporting conventions of LLM-as-judge work \citep{zheng2023judging,dubois2023alpacafarm,liu2023geval}, we report raw agreement together with Cohen's $\kappa$, the sample on which each number was obtained, and the direction of disagreement \citep{agreementmetrics2026}. Table~\ref{tab:judgeval} summarizes.

\paragraph{A1 judge.} On 566 known-answer pairs (156 approved definitions and 410 negative controls), candidate judges achieve 96--97\% binary accuracy. On 300 stratified Sonnet 5 outputs, the judges agree at 79--81\% ($\kappa$ 0.68--0.72), primarily disagreeing at the partial/exact boundary. Claude Opus 5 serves as the judge (GPT-5.6 luna for Claude-family outputs). On a separate 150-judgment sample stratified by judge score across models and regimes, the adopted judge agrees with the human at 81.3\% ($\kappa$ 0.72) over three levels and 89.3\% ($\kappa$ 0.77) for the exact/not-exact outcome used in our tables. Word-only answers have higher agreement (91\%, $\kappa$ 0.85) than usage-conditioned definitions (72\%, $\kappa$ 0.57).

\paragraph{Formation classifier.} On 300 human-labeled items (75 for each of the four frequent mechanisms, compounding, derivation, blending and abbreviation; syntactic formation, 37 lexicon entries, was not benchmarked) Claude Opus 5 reaches 80.3\% ($\kappa$ 0.74) and GPT-5.6 sol 79.7\% ($\kappa$ 0.73) with 89\% mutual agreement ($\kappa$ 0.85); confusions are blend$\leftrightarrow$abbreviation, which merge into 81--82\% when the two classes are pooled. Opus 5 is fixed as classifier for all models; GPT-5.6 sol serves as second annotator. A self-preference check on 462 Claude-generated coinages found Opus 5 four points more generous than GPT-5.6 on mechanism match, within the two classifiers' disagreement on human labels.

\paragraph{G2 judge.} Deterministic checks first: exact labels 240/240, consensus counts 233/240, formation 370/378 against rule-based extraction. Human validation on a stratified sample of 60 sessions (2 exact, 3 near, 19 same-mechanism, 22 different, 14 none by judge label) gives perfect agreement on whether a name formed ($\kappa$ = 1.00), 93\% exact agreement on how many agents shared it, and 98\% ($\kappa$ 0.88) on whether the human coinage was reproduced (exact or near). Separately from the 60-session sample, all 27 sessions judged \emph{near} in the Claude Sonnet 5 ablation runs (four conditions, three seeds) were checked; the human retained 23 as \emph{near} and reassigned 4 to \emph{same mechanism} (85\% precision). Because GPT-5.6 sol is itself an evaluated backbone, we re-judged the main set with Claude Opus 5 (effort low): it reproduces the formation decision for 96.8\% of the 497 sessions, reproduction for 97.0\% and the five-way relation for 85.7\%, with mean formation rates within two points of the reported ones.

\begin{figure}[t!]
    \centering
    \includegraphics[width=\columnwidth]{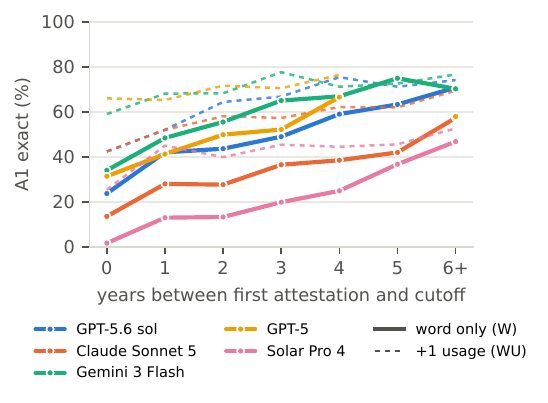}
    \caption{A1 accuracy on pre-cutoff items by age of the word at cutoff (solid: word only; dotted: with one usage).}
    \label{fig:age}
\end{figure}

\section{Acquisition analyses}
\label{app:acq_anal}

\subsection{Pre- and post-cutoff items and exposure}
\label{app:age}
This appendix distinguishes whether a word predates a model's cutoff from how long the word had already been attested by that cutoff. Figure~\ref{fig:age} realigns pre-cutoff items by their age at each model's own cutoff. Table~\ref{tab:seen} gives the number of pre- and post-cutoff items per model and task. When a documented training cutoff is unavailable, the release date is used as the temporal boundary.

\begin{table}[t!]
\centering\scriptsize
\setlength{\tabcolsep}{3pt}
\resizebox{\columnwidth}{!}{%
\begin{tabular}{l rrrrr rrrrrrr}
    \toprule
    & \multicolumn{5}{c}{Neologism reports} & \multicolumn{7}{c}{Web-collected} \\
    \cmidrule(lr){2-6}\cmidrule(lr){7-13}
    Model & 20 & 21 & 22 & 23 & 24 & 20 & 21 & 22 & 23 & 24 & 25 & 26 \\
    \midrule
    GPT-5.6 sol & 71 & 60 & 62 & 46 & 41 & 69 & 64 & 56 & 56 & 57 & 42 & 26 \\
    Claude Sonnet 5 & 59 & 40 & 42 & 35 & 28 & 58 & 42 & 34 & 32 & 30 & 26 & 16 \\
    Gemini 3 Flash & 75 & 65 & 66 & 54 & 48 & 75 & 80 & 74 & 67 & 62 & 35 & 26 \\
    GPT-5 & 67 & 51 & 55 & 40 & 31 & 65 & 58 & 46 & 35 & 28 & 23 & 24 \\
    Solar Pro 4 & 48 & 35 & 31 & 21 & 14 & 42 & 39 & 26 & 21 & 18 & 15 & 5 \\
    GPT-4o & 33 & 22 & 20 & 12 & 8 & 37 & 15 & 7 & 17 & 8 & 3 & 4 \\
    \midrule
    $n$ & 458 & 302 & 394 & 344 & 181 & 100 & 59 & 68 & 72 & 98 & 183 & 153 \\
    \bottomrule
\end{tabular}}
\caption{A1-W accuracy (\%) by first-attestation year for items from the \emph{Neologisms 2020--2024} series and from our web collection; model cutoffs as in Table~\ref{tab:main}.}
\label{tab:age}
\end{table}

\paragraph{Age and provenance.} Table~\ref{tab:seen} reports cutoff-specific item counts and uncertainty; Table~\ref{tab:main} gives the corresponding accuracy. A post-cutoff A1-W success need not imply memorization, because transparent forms can be inferred and effective cutoffs are uncertain. For the five-model temporal analysis, Figure~\ref{fig:age} re-bins pre-cutoff words by age at each cutoff; Table~\ref{tab:age} shows the recency decline separately for expert-report and web-collected words. The pattern persists in both sources and is moderated by semantic transparency, while a usage sentence reduces the age gradient.

\begin{figure}[t!]
    \centering
    \includegraphics[width=\columnwidth]{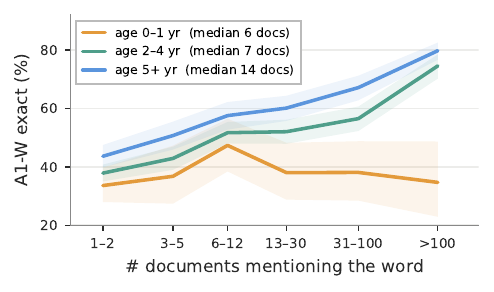}
    \caption{A1-W accuracy on pre-cutoff items by cumulative corpus mentions up to the model's cutoff year (log bins), for three word ages at cutoff; dots = median exposure; five 2025--2026 models pooled.}
    \label{fig:exposure}
\end{figure}

\paragraph{Exposure.} Older words have more observed mentions (median 6, 7 and 14 documents at ages 0--1, 2--4 and 5+), and higher exposure predicts better A1-W accuracy within an age bin (Figure~\ref{fig:exposure}; five 2025--2026 models). In a joint fit, age remains associated with accuracy (+3.5 points per year) after corpus mentions (+5.5 points per log-unit); this corpus-based proxy is not a direct measure of pretraining exposure.

\subsection{Domain and transparency}
\label{app:domain}
\paragraph{Domain labels and circulation.} A model (GPT-5.6 luna, 25 entries per batch) classifies headword and definition alone into institutional ($n$=1,583), popular-cultural (1,302), or colloquial (1,386); these categories describe concepts rather than the genre of occurrence. Restricting to 1,175 entries with at least three corpus mentions and separating news-dominant, mixed and informal-dominant circulation leaves the institutional--colloquial performance gap in place. Hence genre of occurrence alone does not explain the domain effect.

\begin{figure}[t!]
    \centering
    \includegraphics[width=\columnwidth]{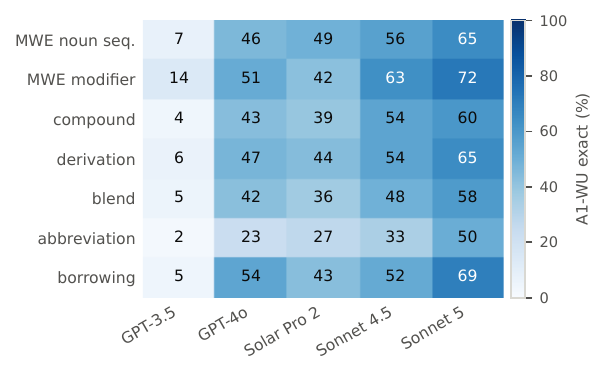}
    \caption{A1-WU exact accuracy by taxonomy leaf. Abbreviations are the hardest leaf for every model; multiword expressions and borrowings the easiest.}
    \label{fig:bytype}
\end{figure}

\paragraph{Formation type.} Figure~\ref{fig:bytype} separates A1-WU by taxonomy leaf. Across proprietary models, abbreviations and blends have the weakest accuracy, while multiword expressions and borrowings are easier; the type ordering weakens within a transparency level. A3 shows a different pattern: with a definition, blends are comparatively easy to analyze (82\% A3-WD for five recent models) and derivations harder (65\%). The one-usage gain is largest for blends (+25 points), consistent with contextual inference helping opaque forms.

\paragraph{Controlling for transparency.} Table~\ref{tab:register} reports A1-W accuracy by domain and transparency. The institutional--colloquial gap decreases substantially among transparent items and differs by formation type, so the aggregate domain effect should not be read as a direct effect of collection genre. The remaining gaps for some models indicate that transparency is not the sole explanation.

\paragraph{Illustrative errors.} A plausible surface segmentation may give the wrong referent: \ko{내세가불} is misread as \ko{내세}+\ko{가불} instead of its source phrase; similarly, an acronym can be captured by an established homograph (\ko{타코트레이드}). These errors help explain why plausible but incorrect glosses receive no A1 exact-match credit.

\begin{table}[t!]
\centering\small
\setlength{\tabcolsep}{4pt}
\resizebox{\columnwidth}{!}{%
\begin{tabular}{l rrr rrr rrr}
    \toprule
    & \multicolumn{3}{c}{High transparency} & \multicolumn{3}{c}{Low transparency} & \multicolumn{3}{c}{Single-word only} \\
    \cmidrule(lr){2-4}\cmidrule(lr){5-7}\cmidrule(lr){8-10}
    \textbf{Model} & Inst. & Pop. & Coll. & Inst. & Pop. & Coll. & Inst. & Pop. & Coll. \\
    \midrule
    GPT-4o & 30 & 33 & 39 & 0 & 0 & 2 & 24 & 22 & 18 \\
    GPT-5 & 74 & 69 & 59 & 21 & 22 & 14 & 55 & 48 & 35 \\
    GPT-5.6 sol & 77 & 72 & 78 & 37 & 27 & 30 & 57 & 52 & 51 \\
    Claude Sonnet 4.5 & 44 & 34 & 35 & 0 & 6 & 2 & 26 & 23 & 16 \\
    Claude Sonnet 5 & 62 & 51 & 57 & 8 & 7 & 5 & 42 & 32 & 25 \\
    Gemini 3 Flash & 83 & 84 & 78 & 55 & 44 & 50 & 63 & 67 & 58 \\
    Solar Pro 4 & 49 & 37 & 36 & 2 & 6 & 5 & 29 & 23 & 19 \\
    HCX-007 & 53 & 50 & 55 & 6 & 10 & 15 & 35 & 36 & 26 \\
    EXAONE 4.5 33B & 31 & 19 & 23 & 4 & 0 & 0 & 15 & 10 & 10 \\
    Qwen3.5 9B & 8 & 4 & 2 & 0 & 1 & 0 & 7 & 4 & 3 \\
    \midrule
    $n$ (max) & 142 & 142 & 92 & 49 & 83 & 73 & 289 & 405 & 490 \\
    \bottomrule
\end{tabular}}
\caption{A1-W accuracy (\%) on pre-cutoff items by domain (Inst.\ = institutional, Pop.\ = popular culture, Coll.\ = colloquial).}
\label{tab:register}
\end{table}

\begin{table}[t!]
\centering\small
\setlength{\tabcolsep}{4pt}
\resizebox{\columnwidth}{!}{%
\begin{tabular}{l r rrr}
    \toprule
    & & \multicolumn{3}{c}{A1 exact (partial)} \\
    \cmidrule(lr){3-5}
    \textbf{Model} & $n$ & W & WU & +search \\
    \midrule
    \rowcolor{gray!15}\multicolumn{5}{l}{\textit{each model's own post-cutoff items}} \\
    GPT-5.6 sol & 53 & 23 (15) & 42 (28) & 70 (23) \\
    Claude Sonnet 5 & 55 & 11 (15) & 42 (27) & 65 (20) \\
    Gemini 3.5 Flash-Lite & 46 & 15 (7) & 17 (26) & 52 (17) \\
    \rowcolor{gray!15}\multicolumn{5}{l}{\textit{items post-cutoff for all three}} \\
    GPT-5.6 sol & 46 & 22 (15) & 41 (28) & 72 (22) \\
    Claude Sonnet 5 & 46 & 11 (15) & 41 (28) & 63 (22) \\
    Gemini 3.5 Flash-Lite & 46 & 15 (7) & 17 (26) & 52 (17) \\
    \bottomrule
\end{tabular}}
\caption{Retrieval ablation on post-cutoff items. Mean queries per five-item request: 3.2 (GPT-5.6 sol), 5.8 (Sonnet 5), 1.8 (Gemini).}
\label{tab:search}
\vspace{-5pt}
\end{table}

\subsection{Retrieval ablation}
\label{app:search}

We compared A1-W with a single usage example and with each vendor's web search on post-cutoff items (Table~\ref{tab:search}). Search substantially improves definition accuracy but does not eliminate the gap: among the 46 items post-cutoff for all three tested models, nine are not defined exactly by any, and two of those are defined by at least one model from a usage sentence. Retrieved glosses and in-context evidence are therefore complementary rather than interchangeable.

\section{Generation analyses}
\subsection{Known words in the generation tasks}
\label{app:knownword}

\begin{table}[t!]
\centering\small
\setlength{\tabcolsep}{3pt}
\resizebox{\columnwidth}{!}{%
\begin{tabular}{l rr rrr}
    \toprule
    & \multicolumn{2}{c}{$n$} & \multicolumn{3}{c}{exact on known items} \\
    \cmidrule(lr){2-3} \cmidrule(lr){4-6}
    \textbf{Model} & unk. & known & M & MD & MDC \\
    \midrule
    \rowcolor{gray!15}\multicolumn{6}{l}{\textit{Proprietary (API)}} \\
    GPT-3.5 Turbo  &  830  &  3  &  --  &  --  &  -- \\
    GPT-4o  &  699  &  86  &  46.5  &  58.1  &  61.5 \\
    GPT-5  &  490  &  277  &  62.8  &  66.1  &  72.6 \\
    GPT-5.4 mini  &  561  &  207  &  64.3  &  60.4  &  69.4 \\
    GPT-5.6 sol  &  355  &  415  &  80.5  &  81.2  &  85.0 \\
    Claude Sonnet 4.5  &  619  &  133  &  54.9  &  57.9  &  68.8 \\
    Claude Sonnet 5  &  549  &  221  &  70.1  &  74.0  &  74.4 \\
    Gemini 3 Flash  &  258  &  474  &  73.8  &  72.4  &  69.1 \\
    Gemini 3.5 Flash-Lite  &  551  &  233  &  60.5  &  57.5  &  64.0 \\
    Solar Pro 2  &  689  &  90  &  45.6  &  51.1  &  52.9 \\
    Solar Pro 4  &  650  &  145  &  40.0  &  51.7  &  50.6 \\
    HCX-003  &  469  &  232  &  44.0  &  49.6  &  46.7 \\
    HCX-007  &  490  &  214  &  48.6  &  48.6  &  50.0 \\
    \midrule
    \rowcolor{gray!15}\multicolumn{6}{l}{\textit{Open-weight}} \\
    Qwen2 7B  &  753  &  17  &  --  &  --  &  -- \\
    Qwen2.5 7B  &  732  &  29  &  24.1  &  13.8  &  -- \\
    Qwen3 8B  &  828  &  3  &  --  &  --  &  -- \\
    Qwen3.5 9B  &  796  &  16  &  --  &  --  &  -- \\
    EXAONE 3.0 7.8B  &  595  &  115  &  18.3  &  22.6  &  11.8 \\
    EXAONE 3.5 7.8B  &  667  &  75  &  24.0  &  20.0  &  20.5 \\
    EXAONE 4.0 32B  &  834  &  3  &  --  &  --  &  -- \\
    EXAONE 4.5 33B  &  735  &  61  &  29.5  &  36.1  &  41.9 \\
    \bottomrule
\end{tabular}}
\caption{G1 exact match (\%) for words already defined correctly in A1-W. Counts of unknown (unk.) and known evaluation items are shown for comparison; results on unknown words are in Table~\ref{tab:main}. Cells with fewer than 20 known items are suppressed. Items receiving partial A1-W credit are excluded from both splits, so counts need not sum to 839.}
\label{tab:g1known}
\end{table}

Both generation tasks are reported on words that the evaluated model could not define from the word alone, i.e.\ items scored wrong or abstained in A1-W (the behavioral \emph{unknown} split of \S\ref{sec:koage}); for G2 the criterion is applied per backbone to the card word, and a mixed community is excluded when any of its backbones knows the word. Table~\ref{tab:g1known} shows why: on words a model already knows, rebuilding the form from its materials is far easier, so pooling known and unknown words would credit recall as coinage. The known columns also serve as a check that the task behaves as designed. Because the unknown-card subset differs by backbone (18--25 cards), cross-model differences may partly reflect variation in evaluated concepts. On the 15 cards unknown to all 21 backbones, the broad pattern persists: six recent proprietary models form names on 93--100\% of cards, whereas open-weight models span 0--60\%. This controls for card composition, though the common subset remains small.

\subsection{G2 ablations and community dynamics}
\label{app:g2abl}

\begin{table}[t!]
\centering\small
\setlength{\tabcolsep}{4pt}
\begin{tabular}{l rr}
    \toprule
    \textbf{Backbone} & Form. & Single-word \\
    \midrule
    \multicolumn{3}{l}{\textit{Proprietary (API)}} \\
    GPT-3.5 Turbo & 0 & 0 \\
    GPT-4o & 20 & 12 \\
    GPT-5 & 86 & 77 \\
    GPT-5.4 mini & 65 & 65 \\
    GPT-5.6 sol & 100 & 89 \\
    Claude Sonnet 4.5 & 95 & 90 \\
    Claude Sonnet 5 & 96 & 96 \\
    Gemini 3 Flash & 100 & 100 \\
    Gemini 3.5 Flash-Lite & 92 & 92 \\
    Solar Pro 2 & 67 & 50 \\
    Solar Pro 4 & 79 & 75 \\
    HCX-003 & 29 & 21 \\
    HCX-007 & 67 & 50 \\
    \multicolumn{3}{l}{\textit{Open-weight}} \\
    Qwen2 7B & 12 & 4 \\
    Qwen2.5 7B & 0 & 0 \\
    Qwen3 8B & 32 & 24 \\
    Qwen3.5 9B & 44 & 24 \\
    EXAONE 3.0 7.8B & 16 & 8 \\
    EXAONE 3.5 7.8B & 60 & 32 \\
    EXAONE 4.0 32B & 4 & 4 \\
    EXAONE 4.5 33B & 24 & 24 \\
    \bottomrule
\end{tabular}
\caption{G2 formation under the judge criterion (label shared by $\geq$2 agents) and under a stricter single-word criterion (no space or punctuation, $\leq$8 characters), in percent. Reproduction is reported in Table~\ref{tab:main}.}
\label{tab:g2strict}
\end{table}

\paragraph{Cross-backbone formation.} Open-weight backbones spread from 0\% to 60\% with no trend that two model sizes can establish; the two 32--33B EXAONE models form names in 4\% and 24\% of sessions. The judge counts a session as formed when at least two agents report the same label at the exit interview; for the weakest open models that label is often a descriptive phrase (7 of EXAONE 3.5 7.8B's 15, 5 of Qwen3.5 9B's 11), so Table~\ref{tab:g2strict} adds a stricter single-word criterion, which reduces formation rates unevenly across backbones, with larger drops for models that often produce descriptive phrases rather than lexicalized forms.

\paragraph{Naming dynamics.} Early naming is common but varies by backbone: the final consensus name emerges in the first episode in 67--84\% of formed sessions for several frontier models versus 29--50\% for others. In the mixed group, GPT-5.6 sol and Gemini 3 Flash propose 25 of 31 shared names (13 and 12, respectively), although the proposer is the first speaker in only 8 sessions. This suggests that proposing and adopting names are different behaviors; the two seeds rotate backbones across personas.

\paragraph{Heterogeneous community.}
\label{app:mixed}

\begin{table}[t!]
\centering\small
\setlength{\tabcolsep}{4pt}
\resizebox{\columnwidth}{!}{%
\begin{tabular}{l rrrr}
    \toprule
    \textbf{Community} & Form. & Repro. & Exact & Abbr. \\
    \midrule
    Mixed (4 backbones, 2 seeds) & 97 & 53 & 25 & 42 \\
    \midrule
    GPT-5.6 sol alone & 100 & 47 & 26 & 42 \\
    Gemini 3 Flash alone & 100 & 44 & 17 & 56 \\
    Claude Sonnet 5 alone & 96 & 29 & 17 & 26 \\
    Solar Pro 4 alone & 79 & 12 & 0 & 26 \\
    \bottomrule
\end{tabular}}
\caption{Mixed vs.\ homogeneous communities under the full protocol (\%; unknown cards only: 32 mixed sessions over two seeds, 18--24 homogeneous sessions). Form.\ = shared name; Repro.\ = exact or near reproduction; Abbr.\ = abbreviations among formed names.}
\label{tab:mix}
\end{table}

The main runs use four agents of one backbone. To test whether coinage depends on community composition, we ran the full protocol on the 25 cards with one agent each from GPT-5.6 sol, Claude Sonnet 5, Gemini 3 Flash and Solar Pro 4, with two seeds that rotate the assignment of backbones to personas (32 sessions on the 16 cards whose word none of the four backbones knew). Table~\ref{tab:mix} compares the mixed community with each backbone alone. The mixture performs at the level of its strongest members, not at their mean, and card-level formation agrees across the two seeds on 15 of 16 cards.



\section{Robustness checks}
\label{app:robust}
This appendix tests the main findings against alternative definitions and controls, using the runs already reported unless stated otherwise.

\paragraph{Temporal split (Finding~2).} First-attestation years are coarser than cutoff months, and the release month stands in for an unpublished cutoff. Table~\ref{tab:robust_time} tightens the post-cutoff definition for models with a documented cutoff: items first attested at least one year after the cutoff, the recorded-basis subset of those, and month-dated items strictly after the cutoff month. Word-only recall stays at 10--13\% under every definition against 34.8\% pre-cutoff, and the cutoff-year band (14.1\%) sits between the two, so neither dating granularity nor cutoff-adjacent items drive the gap. Models with a 2026 cutoff have no item a full year after it yet. Post-cutoff items that a model nevertheless defines are concentrated in transparent forms: 19.5\% of high-transparency post-cutoff items are defined against 4.8\% of mid- and 1.9\% of low-transparency ones (proprietary pooled), which points to compositional inference rather than leakage.
\begin{table}[t!]
\centering\small
\setlength{\tabcolsep}{3pt}
\resizebox{\columnwidth}{!}{%
\begin{tabular}{l c rrrrrr}
\toprule
\textbf{Model} & \textbf{Cutoff} & pre & post & $\geq$cut.+1 & recorded & month & band \\
\midrule
GPT-3.5 Turbo & 2021-09 & 0.9 & 0.1 & 0.1 & 0.2 & 0.2 & 0.5 \\
GPT-4o & 2023-10 & 22.1 & 7.1 & 5.8 & 7.8 & 7.3 & 12.6 \\
GPT-5 & 2024-09 & 51.2 & 21.2 & 20.4 & 23.9 & 21.3 & 29.9 \\
GPT-5.4 mini & 2025-08 & 32.1 & 15.4 & 12.6 & 16.7 & 15.4 & 15.7 \\
GPT-5.6 sol & 2026-02 & 57.5 & 19.5 & -- & -- & 19.5 & 22.1 \\
Claude Sonnet 4.5 & 2025-07 & 30.0 & 10.8 & 7.4 & 11.1 & 10.8 & 15.7 \\
Claude Sonnet 5 & 2026-01 & 40.8 & 11.2 & -- & -- & 11.2 & 12.6 \\
Gemini 3 Flash & 2025-01 & 64.9 & 25.8 & 21.1 & 22.2 & 25.8 & 30.0 \\
Gemini 3.5 Flash-Lite & 2026-03 & 32.9 & 12.5 & -- & -- & 12.5 & 11.6 \\
Solar Pro 2 & 2025-03 & 20.4 & 2.1 & 3.2 & 11.1 & 2.1 & 2.9 \\
Solar Pro 4 & 2026-02 & 30.0 & 1.2 & -- & -- & 1.2 & 1.1 \\
EXAONE 4.5 33B & 2024-12 & 17.8 & 1.7 & 1.7 & 1.4 & 1.9 & 4.2 \\
Mean (proprietary) & & 34.8 & 11.5 & 10.1 & 13.3 & 11.6 & 14.1 \\
\bottomrule
\end{tabular}}
\caption{Temporal sensitivity of A1-W (exact, \%) for models with a documented cutoff: pre-cutoff items; all post-cutoff items; items first attested at least one year after the cutoff; the recorded-basis subset of those; month-dated items strictly after the cutoff month; and items from the cutoff year (band). ``--'': no item yet.}
\label{tab:robust_time}
\end{table}

\paragraph{Contextual choice (Finding~1).} A2 could be solved by elimination rather than by acquiring the meaning. Table~\ref{tab:robust_a2} conditions A2 on the model's own A1-W outcome: on words the model could not define, proprietary accuracy is 85.9\% (96.0\% on words it could define), and errors split between the same-formation-type neologism (42.9\%) and the near-synonym (48.1\%), with the funnel hard negative chosen in only 8.9\%. A2 and A1-WU share the usage sentence for 95\% of A2 items; on the 57 items whose sentences differ, A2 accuracy on unknown words is 89.4\%. On the 1,225 items shared by A1-WU and A2, proprietary means are 50.3\% exact and 80.6\% exact-or-partial for A1-WU against 90.4\% for A2 (37.5, 70.5 and 86.2\% on A1-W-unknown items). Because A2 only shows that the model can place the word, the generative evidence of acquisition is A1-WU: one usage lifts exact definitions of unknown words to 39.8\% pre-cutoff and 25.8\% post-cutoff.
\begin{table}[t!]
\centering\small
\setlength{\tabcolsep}{3pt}
\resizebox{\columnwidth}{!}{%
\begin{tabular}{l rrr rrr}
\toprule
& \multicolumn{3}{c}{A2 accuracy (\%)} & \multicolumn{3}{c}{Distractor chosen in errors (\%)} \\
\cmidrule(lr){2-4}\cmidrule(lr){5-7}
\textbf{Model} & all & unknown & known & same type & hard neg. & near syn. \\
\midrule
GPT-3.5 Turbo & 66.1 & 65.7 & 80.0 & 33.3 & 14.9 & 51.8 \\
GPT-4o & 89.4 & 85.5 & 94.8 & 46.2 & 10.0 & 43.8 \\
GPT-5 & 96.7 & 92.5 & 99.5 & 40.0 & 2.5 & 57.5 \\
GPT-5.4 mini & 89.2 & 83.4 & 95.0 & 42.9 & 7.5 & 49.6 \\
GPT-5.6 sol & 97.9 & 95.0 & 99.0 & 53.8 & 7.7 & 38.5 \\
Claude Sonnet 4.5 & 90.7 & 85.4 & 98.4 & 52.6 & 11.4 & 36.0 \\
Claude Sonnet 5 & 92.7 & 85.8 & 98.8 & 33.3 & 8.9 & 57.8 \\
Gemini 3 Flash & 98.1 & 96.3 & 98.6 & 43.5 & 8.7 & 47.8 \\
Gemini 3.5 Flash-Lite & 94.9 & 92.7 & 98.1 & 46.0 & 3.2 & 50.8 \\
Solar Pro 2 & 88.8 & 84.7 & 97.5 & 50.7 & 13.0 & 36.2 \\
Solar Pro 4 & 92.4 & 88.5 & 97.6 & 35.5 & 15.1 & 49.5 \\
HCX-003 & 87.6 & 79.0 & 95.4 & 40.8 & 5.3 & 53.9 \\
HCX-007 & 89.3 & 82.0 & 94.7 & 39.4 & 7.6 & 52.3 \\
Mean (proprietary) & 90.3 & 85.9 & 96.0 & 42.9 & 8.9 & 48.1 \\
\midrule
Qwen2 7B & 70.9 & 66.2 & 92.2 & 40.2 & 10.3 & 45.8 \\
Qwen2.5 7B & 72.3 & 65.7 & 90.6 & 34.6 & 12.6 & 52.2 \\
Qwen3 8B & 75.3 & 73.2 & 96.8 & 35.9 & 12.8 & 51.3 \\
Qwen3.5 9B & 81.5 & 80.0 & 96.4 & 52.4 & 17.2 & 30.4 \\
EXAONE 3.0 7.8B & 69.8 & 56.4 & 88.6 & 29.1 & 11.3 & 59.6 \\
EXAONE 3.5 7.8B & 75.5 & 66.0 & 91.9 & 30.6 & 9.0 & 60.5 \\
EXAONE 4.0 32B & 80.3 & 80.0 & 100.0 & 43.8 & 12.4 & 43.8 \\
EXAONE 4.5 33B & 79.5 & 72.6 & 94.9 & 29.4 & 6.3 & 64.3 \\
Mean (open-weight) & 75.7 & 70.0 & 93.9 & 37.0 & 11.5 & 51.0 \\
\bottomrule
\end{tabular}}
\caption{A2 accuracy on all items and conditional on A1-W outcome (unknown = wrong or abstained; known = exact), and which distractor the model chose when wrong (share of errors; unparsed answers omitted).}
\label{tab:robust_a2}
\end{table}

\paragraph{Formation analysis (Finding~3).} Table~\ref{tab:robust_a3} replaces the aggregate gain with item-level transitions. Adding the definition turns 12.0\% of pre-cutoff items from wrong to right but 8.6\% from right to wrong, and 34.2\% stay wrong; post-cutoff the gain rises to 18.0\% against a loss of 7.2\%. The gain grows as the word gets younger at the cutoff (11.5\% at three or more years, 13.3\% at one to two, 14.3\% at zero, 14.3\% post-cutoff; pooled proprietary) and is largest for blends and abbreviations post-cutoff (+11.7 and +11.9 points). Reweighting post-cutoff items to each model's pre-cutoff formation-type mix leaves the WD gap at $-1.6$ points (observed $-2.1$), so composition does not explain the reversal. Conditioning on the A1-W outcome instead, the definition adds 5.3 points on pre-cutoff words the model could not define but 0.2 points on words it defined correctly (13 proprietary models), the pattern the finding predicts.

As a further intervention, we gave the gold constituents together with the definition (WDM; Table~\ref{tab:robust_wdm}). For GPT-5.6 sol, Claude Sonnet 5 and Gemini 3 Flash this corrects only 20--31\% of the pre-cutoff items that stay wrong after the definition, and for Claude Sonnet 5 it lowers accuracy below WD; 9\% of pre-cutoff items remain wrong under all three regimes, concentrated at two boundaries, derivation read as compounding and abbreviation read as blending, where human annotators also disagree. Segmentation failure therefore does not explain the residual, but part of it is category ambiguity rather than a stored misanalysis, which is why we describe the prior as resistant to the definition rather than as hallucinated.

\begin{table}[t!]
\centering\small
\setlength{\tabcolsep}{3pt}
\resizebox{\columnwidth}{!}{%
\begin{tabular}{l rrr rrr}
\toprule
& \multicolumn{3}{c}{pre-cutoff} & \multicolumn{3}{c}{post-cutoff} \\
\cmidrule(lr){2-4}\cmidrule(lr){5-7}
\textbf{Model} & gain & loss & stay wrong & gain & loss & stay wrong \\
\midrule
GPT-3.5 Turbo & 9.6 & 12.8 & 66.7 & 10.7 & 13.0 & 68.0 \\
GPT-4o & 13.8 & 9.7 & 34.6 & 16.9 & 9.4 & 34.2 \\
GPT-5 & 14.5 & 7.9 & 17.6 & 23.1 & 4.5 & 17.9 \\
GPT-5.4 mini & 15.4 & 10.4 & 23.6 & 19.2 & 5.1 & 24.2 \\
GPT-5.6 sol & 8.4 & 5.5 & 12.1 & 17.3 & 1.9 & 15.4 \\
Claude Sonnet 4.5 & 13.0 & 8.8 & 25.6 & 19.2 & 3.8 & 25.0 \\
Claude Sonnet 5 & 15.9 & 6.2 & 16.0 & 25.5 & 5.5 & 16.4 \\
Gemini 3 Flash & 7.0 & 5.4 & 17.0 & 10.8 & 5.8 & 21.6 \\
Gemini 3.5 Flash-Lite & 10.7 & 6.4 & 27.5 & 17.8 & 13.3 & 20.0 \\
Solar Pro 2 & 16.3 & 11.9 & 39.4 & 18.9 & 11.0 & 38.6 \\
Solar Pro 4 & 16.8 & 13.0 & 30.1 & 36.5 & 1.9 & 25.0 \\
HCX-003 & 5.3 & 5.8 & 71.8 & 5.2 & 8.9 & 68.5 \\
HCX-007 & 9.0 & 8.6 & 61.9 & 12.5 & 9.6 & 60.6 \\
\midrule
Qwen2 7B & 12.9 & 8.4 & 68.1 & 10.8 & 7.6 & 64.3 \\
Qwen2.5 7B & 8.3 & 10.9 & 66.8 & 9.0 & 11.5 & 64.1 \\
Qwen3 8B & 9.2 & 12.1 & 60.9 & 12.8 & 8.8 & 54.4 \\
Qwen3.5 9B & 13.1 & 10.6 & 58.3 & 23.1 & 7.7 & 50.0 \\
EXAONE 3.0 7.8B & 6.0 & 8.5 & 70.5 & 12.1 & 8.9 & 63.7 \\
EXAONE 3.5 7.8B & 6.2 & 10.5 & 73.0 & 9.2 & 6.5 & 68.6 \\
EXAONE 4.0 32B & 9.1 & 9.3 & 64.0 & 11.5 & 8.7 & 61.5 \\
EXAONE 4.5 33B & 9.9 & 15.9 & 51.6 & 13.1 & 9.8 & 54.9 \\
\midrule
Mean (proprietary) & 12.0 & 8.6 & 34.2 & 18.0 & 7.2 & 33.5 \\
\bottomrule
\end{tabular}}
\caption{A3 item-level transitions when the definition is added (W $\rightarrow$ WD): share of items that become correct (gain), become wrong (loss), or stay wrong, pre- vs post-cutoff. Means are over the 13 proprietary models.}
\label{tab:robust_a3}
\end{table}

\begin{table}[t!]
\centering\small
\setlength{\tabcolsep}{3pt}
\resizebox{\columnwidth}{!}{%
\begin{tabular}{l rrrr rrrr}
\toprule
& \multicolumn{4}{c}{pre-cutoff} & \multicolumn{4}{c}{post-cutoff} \\
\cmidrule(lr){2-5}\cmidrule(lr){6-9}
\textbf{Model} & W & WD & WDM & fixed & W & WD & WDM & fixed \\
\midrule
GPT-5.6 sol & 83.0 & 85.7 & 86.6 & 20 & 76.7 & 88.4 & 86.0 & 25 \\
Claude Sonnet 5 & 71.0 & 80.9 & 70.6 & 27 & 62.2 & 84.4 & 73.3 & 0 \\
Gemini 3 Flash & 78.6 & 80.2 & 83.6 & 31 & 73.7 & 78.9 & 84.2 & 39 \\
Solar Pro 4 & 55.1 & 57.8 & 61.2 & 32 & 39.5 & 74.4 & 65.1 & 30 \\
\bottomrule
\end{tabular}}
\caption{A3 with gold constituents added to definition (WDM) on the 1,203 FORM items that have materials: accuracy (\%) under W, WD and WDM, and the share of items still wrong after WD that WDM corrects (fixed).}
\label{tab:robust_wdm}
\vspace{-6pt}
\end{table}

\paragraph{Community coinage (Findings~5--6).} Table~\ref{tab:robust_g2} reports the main G2 set under stricter criteria and controls. Counting a session as formed only when three or all four agents report the same name lowers the proprietary mean from 68.9\% to 61.9\% and 49.8\% but leaves the frontier backbones near ceiling (GPT-5.6 sol 100/100/94.7\%, Claude Sonnet 5 95.8/95.8/87.5\%). The shared name is not an artifact of the exit interview: in 97.2\% of formed proprietary sessions at least two agents had already used it in the chat, and in 88.1\% it recurs in three or more later episodes. Without the two devices (condition A0) formation is 8--29\% for Claude Sonnet 5 over three seeds, 8.3\% for Gemini 3.5 Flash-Lite and 24.0\% for GPT-5.6 terra, which is the no-expectation control; the three Sonnet 5 seeds agree on the card-level outcome for 20 of 24 cards under the full protocol. We therefore describe G2 outcomes as a shared name at the end of a short interaction, not as conventionalization.

\begin{table*}[t!]
\centering\small
\setlength{\tabcolsep}{4pt}
\resizebox{\textwidth}{!}{%
\begin{tabular}{l r rrr rr rr rr r}
\toprule
& & \multicolumn{3}{c}{Formed (\%) by agents agreeing} & \multicolumn{2}{c}{Persistence (\% of formed)} & \multicolumn{2}{c}{Mech.\ agreement $|$ formed} & \multicolumn{2}{c}{G1-M mech.\ match} & \\
\cmidrule(lr){3-5}\cmidrule(lr){6-7}\cmidrule(lr){8-9}\cmidrule(lr){10-11}
\textbf{Backbone} & cards & $\geq$2 & $\geq$3 & 4 & $\geq$2 agents reuse & $\geq$3 later ep. & \% & 95\% CI & \% & 95\% CI & concept-only \\
\midrule
GPT-3.5 Turbo & 25 & 0.0 & 0.0 & 0.0 & -- & -- & -- & -- & 54.5 & 32--77 & 52 \\
GPT-4o & 25 & 20.0 & 12.0 & 4.0 & 100.0 & 100.0 & 60.0 & 20--100 & 72.7 & 55--91 & 36 \\
GPT-5 & 22 & 86.4 & 68.2 & 54.5 & 100.0 & 89.5 & 52.6 & 32--74 & 86.4 & 73--100 & 45 \\
GPT-5.4 mini & 23 & 65.2 & 47.8 & 34.8 & 100.0 & 86.7 & 60.0 & 33--80 & 81.8 & 64--95 & 39 \\
GPT-5.6 sol & 19 & 100.0 & 100.0 & 94.7 & 100.0 & 100.0 & 68.4 & 47--89 & 81.8 & 64--95 & 58 \\
Claude Sonnet 4.5 & 21 & 95.2 & 95.2 & 95.2 & 100.0 & 100.0 & 55.0 & 30--75 & 77.3 & 59--95 & 38 \\
Claude Sonnet 5 & 24 & 95.8 & 95.8 & 87.5 & 100.0 & 100.0 & 65.2 & 43--83 & 81.8 & 64--95 & 46 \\
Gemini 3 Flash & 18 & 100.0 & 94.4 & 88.9 & 100.0 & 100.0 & 61.1 & 39--83 & 90.9 & 77--100 & 17 \\
Gemini 3.5 Flash-Lite & 24 & 91.7 & 87.5 & 58.3 & 95.5 & 81.8 & 45.5 & 27--64 & 90.9 & 77--100 & 42 \\
Solar Pro 2 & 24 & 66.7 & 62.5 & 41.7 & 100.0 & 87.5 & 43.8 & 19--69 & 86.4 & 73--100 & 42 \\
Solar Pro 4 & 24 & 79.2 & 70.8 & 45.8 & 100.0 & 100.0 & 52.6 & 32--74 & 90.9 & 77--100 & 54 \\
HCX-003 & 24 & 29.2 & 16.7 & 4.2 & 71.4 & 42.9 & 28.6 & 0--71 & 72.7 & 55--91 & 46 \\
HCX-007 & 24 & 66.7 & 54.2 & 37.5 & 100.0 & 68.8 & 62.5 & 38--88 & 77.3 & 59--95 & 46 \\
\midrule
Qwen2 7B & 25 & 12.0 & 4.0 & 0.0 & 33.3 & 0.0 & 33.3 & 0--100 & 50.0 & 27--73 & 40 \\
Qwen2.5 7B & 25 & 0.0 & 0.0 & 0.0 & -- & -- & -- & -- & 63.6 & 41--82 & 36 \\
Qwen3 8B & 25 & 32.0 & 16.0 & 4.0 & 75.0 & 75.0 & 62.5 & 25--88 & 72.7 & 55--91 & 48 \\
Qwen3.5 9B & 25 & 44.0 & 32.0 & 24.0 & 100.0 & 81.8 & 45.5 & 18--73 & 59.1 & 36--77 & 48 \\
EXAONE 3.0 7.8B & 25 & 16.0 & 4.0 & 0.0 & 75.0 & 50.0 & 50.0 & 0--100 & 86.4 & 73--100 & 48 \\
EXAONE 3.5 7.8B & 25 & 60.0 & 32.0 & 12.0 & 80.0 & 60.0 & 40.0 & 20--67 & 77.3 & 59--95 & 48 \\
EXAONE 4.0 32B & 25 & 4.0 & 0.0 & 0.0 & 100.0 & 100.0 & 0.0 & 0--0 & 86.4 & 73--100 & 40 \\
EXAONE 4.5 33B & 25 & 24.0 & 12.0 & 12.0 & 100.0 & 100.0 & 33.3 & 0--67 & 100.0 & 100--100 & 48 \\
\bottomrule
\end{tabular}}
\caption{G2 robustness on the main set (full protocol, one seed, cards unknown to the backbone). Formation under stricter consensus thresholds; persistence of the shared name in the chat log (reused by at least two agents before the exit interview; reappearing in at least three later episodes); mechanism agreement with the human coinage among formed sessions with a session-level bootstrap 95\% interval; G1-M mechanism match on the same words (22 of the 25 are G1 items); and mechanism match of a single-model concept-only coinage control (same concept text, no materials, no community).}
\label{tab:robust_g2}
\end{table*}

\begin{table}[t!]
\centering\small
\setlength{\tabcolsep}{3pt}
\resizebox{\columnwidth}{!}{%
\begin{tabular}{l r rr rr}
\toprule
& & \multicolumn{2}{c}{A1-W} & \multicolumn{2}{c}{A1-WU} \\
\cmidrule(lr){3-4}\cmidrule(lr){5-6}
\textbf{Slice} & items & prop. & open & prop. & open \\
\midrule
\rowcolor{gray!15}\multicolumn{6}{l}{\textit{Provenance}} \\
\quad Neologism reports & 1671 & 35.7 & 12.3 & 51.4 & 31.6 \\
\quad Web API track & 569 & 31.6 & 7.3 & 45.2 & 26.5 \\
\quad Online corpus track & 162 & 37.3 & 9.2 & 59.4 & 33.2 \\
\rowcolor{gray!15}\multicolumn{6}{l}{\textit{Formation type}} \\
\quad FORM-CPD & 223 & 36.0 & 7.6 & 51.5 & 28.5 \\
\quad FORM-DER & 198 & 39.4 & 14.9 & 55.5 & 33.7 \\
\quad FORM-BLD & 425 & 28.0 & 6.2 & 50.6 & 26.2 \\
\quad FORM-ABB & 341 & 23.1 & 3.1 & 41.3 & 14.0 \\
\quad MWE-NSEQ & 969 & 39.3 & 16.3 & 53.4 & 37.5 \\
\quad MWE-MOD & 70 & 45.0 & 14.5 & 61.0 & 48.0 \\
\quad MWE-CLS & 23 & 49.7 & 10.9 & -- & -- \\
\quad BORROW & 159 & 36.5 & 7.6 & 57.1 & 33.1 \\
\rowcolor{gray!15}\multicolumn{6}{l}{\textit{Verification confidence (web)}} \\
\quad confidence $<$0.75 & 339 & 27.2 & 6.1 & 46.7 & 29.0 \\
\quad 0.75--0.9 & 206 & 36.5 & 8.7 & 44.0 & 22.2 \\
\quad $\geq$0.9 & 11 & 73.2 & 15.9 & -- & -- \\
\rowcolor{gray!15}\multicolumn{6}{l}{\textit{Usages in the lexicon}} \\
\quad no usage & 420 & 33.5 & 7.5 & -- & -- \\
\quad 1--2 usages & 1123 & 27.5 & 8.9 & 48.1 & 29.3 \\
\quad 3+ usages & 889 & 43.9 & 15.1 & 55.4 & 33.7 \\
\bottomrule
\end{tabular}}
\caption{A1 accuracy (\%, pre-cutoff items; proprietary and open-weight means) by lexicon provenance, formation type, verification confidence of web-collected entries, and number of attested usages in the lexicon. Slices with fewer than 20 items are omitted.}
\label{tab:robust_slices}
\end{table}

\paragraph{Single versus community coinage (Finding~5).} G1 supplies the materials of the attested word while G2 does not, so part of the gap in mechanism agreement could reflect the input rather than a preference. Table~\ref{tab:robust_g2} gives mechanism agreement conditional on formation with session-level bootstrap intervals (proprietary 29--68\%, e.g.\ GPT-5.6 sol 68.4\% [47--89]), G1-M mechanism match on the same 22 words (55--91\%, e.g.\ Gemini 3 Flash 90.9\% [77--100]), and a control in which a single model receives only the card's concept description, with no materials and no community, and coins one word. In this control the proprietary mean mechanism match is 43.1\% (open-weight 44.5\%), close to the community figure and far below G1-M, so most of the single-versus-community gap in Finding~5 reflects the materials given to G1 rather than a preference of the lone model. The community's distinctive effect survives the control: abbreviations make up 32\% of community names against 7.2\% of concept-only coinages and 16\% of the human words, while concept-only coinage favors blends (45\% of outputs). Figure~\ref{fig:sankey} pools formed sessions only, so its comparison is conditional on a name having formed.

\paragraph{Coverage and reliability (Finding~4).} Table~\ref{tab:robust_slices} splits pre-cutoff A1 accuracy by lexicon provenance, formation type, verification confidence of web-collected entries, and the number of attested usages. Accuracy is similar across the three tracks (A1-W 31.6--37.3\%), lowest for abbreviations and blends, and rises with verification confidence and with the number of usages, which the lexicon reports per entry. The judged pool of 2,000 candidates is chosen for precision, so the lexicon under-represents rare and weakly attested coinages; a second native-speaker annotator independently re-labeled the precision sample (212 items: 100 online-corpus, 97 web, 15 pipeline-rejected controls) and the 150-item A1 judge sample: the two annotators agree on 94.8\% of precision decisions (online corpus 94\% vs.\ 92\% accepted, web 98\% vs.\ 99\%) and on 84.0\% of A1 scores ($\kappa$ 0.75), and the A1 judge agrees with the second annotator at 81.3\% ($\kappa$ 0.72), the same as with the first. On the 60 G2 validation sessions the second annotator matched the first on every formation, count and relation decision and on 58 of 60 mechanism labels; the 300 mechanism labels of the classifier benchmark are themselves human-consensus labels.

\section{Prompts}
\label{app:prompts}
The Korean evaluation prompt templates are reproduced below, with placeholders for item-specific inputs and English translations in italics. Acquisition and G1 requests contain ten items in one user turn with JSON output; G2 follows the multi-turn community protocol of \S\ref{sec:koage}.

\vspace{\baselineskip}

\begin{promptbox}{A1 definition, word only (W)}
\ko{각 한국어 단어의 뜻을 한 문장으로 정의하세요. 모르는 단어면 추측하지 말고 definition에 "모름"이라고 쓰세요.}\\
\ko{1. '단어'} \ldots\\
\ko{JSON 배열로만 출력: [\{"idx": 1, "definition": "..."\}]}\\[2pt]
\textit{Define the meaning of each Korean word in one sentence. If you do not know the word, do not guess; write ``don't know'' in \texttt{definition}. Output only a JSON array.}
\end{promptbox}

\begin{promptbox}{A1 definition, with one usage (WU)}
\ko{각 한국어 단어의 뜻을 제시된 용례를 참고하여 한 문장으로 정의하세요. 용례만으로 뜻을 알 수 없으면 definition에 "모름"이라고 쓰세요.}\\
\ko{1. '단어'} \quad \ko{용례: \ldots}\\[2pt]
\textit{Define each word in one sentence using the given usage sentence; if the usage does not reveal the meaning, write ``don't know''.}
\end{promptbox}

\begin{promptbox}{A2 contextual choice}
\ko{각 문장의 [MASKED] 자리에 가장 알맞은 말을 선택지에서 고르세요. 선택지 기호만 답하세요.}\\
\ko{1. \ldots [MASKED] \ldots \quad (A) \ldots / (B) \ldots / (C) \ldots / (D) \ldots}\\[2pt]
\textit{Choose the option that best fills [MASKED] in each sentence; answer with the letter only.} Options are the answer, a same-formation-type neologism, a hard negative from the verification funnel and a near-synonym, in an item-seeded random order.
\end{promptbox}

\begin{promptbox}{A3 formation analysis (W / WD)}
\ko{각 신어가 어떤 형태적 기제로 만들어졌는지 선택지에서 하나 고르세요. 선택지 기호만 답하세요.}\\[2pt]
\textit{Choose how each neologism was formed; answer with the letter only.} The item shows the word (W) or the word with its definition (WD) and five options naming the mechanisms (compounding, derivation, blending, abbreviation, syntactic fusion) with the decomposition proposed for each, plus ``not decomposable''.
\end{promptbox}

\begin{promptbox}{G1 coinage (M / MD / MDC)}
\ko{당신은 한국어 신조어를 만드는 언중의 한 사람입니다. 각 항목마다 주어진 **재료**만을 바탕으로 \{새 한국어 단어(재료가 가리킬 만한 개념을 스스로 정해) | **개념**을 나타내는 새 한국어 단어\}를 하나 만드세요.}\\
\ko{- 재료를 그대로 이어 붙여도 되고, 일부 음절만 잘라 합치거나(혼성), 첫 음절들로 줄이거나(두음절어), 접사처럼 붙여도 됩니다. 실제 한국어 화자가 만들 법한 자연스러운 형태여야 합니다.}\\
\ko{- 재료에 없는 말을 새로 끌어오지 마세요(조사·접사 정도는 허용).}\\
\ko{- 각 항목에 대해 가장 그럴듯한 단어 1개와 대안 2개를 제시하세요. 설명은 쓰지 마세요.}\\
\ko{항목: 1. 재료: \ldots [개념: \ldots] [쓰임 예(\_\_\_ 자리에 들어갈 단어): \ldots]}\\[2pt]
\textit{You are one of the speakers who coin Korean neologisms. From the given materials only, make one new Korean word \{for a concept you choose | for the given concept\}. You may join the materials as they are, cut and combine syllables (blending), take initial syllables (initialism), or attach them like affixes; the form must be one a real speaker would produce. Do not bring in material that is not given (particles and affixes are allowed). Give the most plausible word and two alternatives; no explanation.} Only the most plausible word is scored. Items give the materials (M), plus the concept (MD), plus a masked usage context (MDC).
\end{promptbox}

\begin{promptbox}{G2 agent rules (first user turn, every agent)}
\ko{[시뮬레이션 안내] 이것은 모바일 채팅방 커뮤니티 역할극이다. 너는 \{community\} 멤버 '\{name\}' 역할이다. \{persona\} \{culture\} 답장은 모바일 채팅답게 짧게 쓴다.}\\
\ko{\{concept\_line\} (이것을 부르는 정해진 말은 아직 없다.)}\\
\ko{진행자가 상황을 주면 그 상황의 대화에 자연스럽게 참여하면 된다. 자기 차례에 채팅 1~2개만 보내라. 평소 말투로 자연스럽게.}\\[2pt]
\textit{[Simulation notice] This is a community role-play in a mobile group chat. You are `\{name\}', a member of \{community\}. \{persona\} \{culture\} Write short mobile-chat replies. \{concept line\} (There is no fixed word for this yet.) When the host gives a situation, join the conversation naturally; send one or two chat messages per turn in your usual voice.}
\end{promptbox}

\begin{promptbox}{G2 method sentence (appended to the culture text)}
Full protocol: \ko{참고로 한국어에서는 자주 쓰는 긴 표현을 새 말로 만들 수 있다 --- 단어를 그대로 이어 붙이거나(합성), 접사를 붙이거나(파생), 단어 조각을 붙이거나(혼성), 첫 음절들로 줄이는(축약) 방식이 흔하다.}\\
\textit{Korean speakers can make new words for frequent long expressions: joining words (compounding), attaching affixes (derivation), joining fragments (blending) or taking initial syllables (abbreviation) are common.}\\[3pt]
Shortening-only variant: \ko{참고로 한국어에서는 자주 쓰는 긴 표현을 새 말로 줄여 만들 수 있다 --- 각 단어의 첫 음절을 따거나(두음절어), 단어 조각을 붙이거나(혼성), 앞부분만 남기는(절단) 방식이 흔하다.}\\
\textit{Korean speakers can shorten frequent long expressions into new words: taking each word's first syllable (initialism), joining fragments (blending) or keeping only the front (clipping) are common.}\\[3pt]
The context-only and no-devices conditions carry no method sentence; the no-devices condition also removes personas and culture.
\end{promptbox}

\begin{promptbox}{G2 host line (example) and exit interview}
Host line: \ko{박람회·직영 숍에서 잘 팔린 불교풍 상품과 젊은 손님 반응을 공유하라.}\\
\textit{Share which Buddhist-style goods sold well at the fair and the shop, and how young customers reacted.}\\[3pt]
Exit interview (each agent, in isolation): \ko{게임이 끝났다. 이번 게임에서 '\{definition\}' 상황을 부르는 말이 생겼나? 뭐라고 불렀고 뜻은 무엇인가? 생기지 않았다면 '없었다'라고 답하라.}\\
\textit{The game is over. In this game, did a word arise for `\{concept\}'? What was it called and what does it mean? If none arose, answer `none'.}
\end{promptbox}

\paragraph{Judge prompts.} The judge instructions are longer and are released with the code rather than reproduced here; their inputs and outputs are as follows. The A1 judge receives the word, the reference definition and the model definition and returns 2 (same core referent and use), 1 (related but different scope, focus or referent) or 0 (different meaning, hallucination, literal gloss or old sense). The G1 mechanism classifier receives the materials and the produced word and returns one of the five FORM mechanisms or \texttt{NONE}. The G2 judge receives the four exit-interview answers and the human coinage with its mechanism and returns the consensus name, the number of agents sharing it, its relation to the human word (exact / near / same mechanism / different / none) and its mechanism. The verification judge of the collection pipeline follows the rules summarized in Appendix~\ref{app:collection}.

\end{CJK}
\end{document}